\DocumentMetadata{}
\documentclass[sigconf, screen, authorversion]{acmart}
\AtBeginDocument{%
  \providecommand\BibTeX{{%
    \normalfont B\kern-0.5em{\scshape i\kern-0.25em b}\kern-0.8em\TeX}}}

\setcopyright{acmlicensed}
\copyrightyear{2024}
\acmYear{2024}
\acmDOI{XXXXXXX.XXXXXXX}

\acmConference[Preprint]{Preprint}{2024}{}
\acmISBN{978-1-4503-XXXX-X/18/06}

\usepackage{multirow}
\usepackage{fancyvrb}
\usepackage{amsmath}
\usepackage{fontawesome5}
\usepackage[htt]{hyphenat}
\usepackage{listings}
\usepackage{xcolor}%
\usepackage{wrapfig}
\usepackage[vlined,linesnumbered]{algorithm2e}
\usepackage{enumitem}
\usepackage{subfigure}
\usepackage{subcaption}
\usepackage{caption}
\usepackage{graphicx}

\usepackage{booktabs}
\usepackage{colortbl}
\usepackage{array}

\definecolor{lightgray}{gray}{0.95}

\definecolor{codebg}{rgb}{0.96,0.96,0.96}
\definecolor{codeframe}{rgb}{0.85,0.85,0.85}

\lstdefinestyle{customc}{
  backgroundcolor=\color{codebg},       % background color
  breaklines=true,                      % enable line breaking
  frame=single,                         % frame style
  rulecolor=\color{codeframe},          % frame color
  xleftmargin=\parindent,               % left margin
  language=Bash,                        % language (bash in this case)
  showstringspaces=false,               % do not show spaces in strings as special characters
  basicstyle=\footnotesize\ttfamily,    % basic style
  keywordstyle=\text{},   % keyword style
  commentstyle=\itshape\color{purple},  % comment style
  identifierstyle=\color{black},        % identifier style
  stringstyle=\color{orange},           % string literal style
  numbers=none,                         % line numbers style
  mathescape=true,                      % allows escaping to math mode in $
  escapechar=@                          % escape character for LaTeX within the listing
}

\newcommand{\systemname}[1]{\textit{Interactive-Reflective Dialogue Alignment}}

\begin{document}

% \title[Democratizing Reward Design]{Democratizing Reward Design for Personal Value-Alignment}

\title[Democratizing Reward Design]{Democratizing Reward Design for Personal and Representative Value-Alignment}

% \title[Reflect, Design, Align, Repeat]{Reflect, Design, Align, Repeat: Accessible and Personalized Reward Design for Value-Alignment}

\author{Carter Blair}
\affiliation{%
  \institution{University of Waterloo}
  \city{Waterloo}
  \country{Canada}}
\email{cblair@uwaterloo.ca}

\author{Kate Larson}
\affiliation{%
  \institution{University of Waterloo}
  \city{Waterloo}
  \country{Canada}}
\email{kate.larson@uwaterloo.ca}

\author{Edith Law}
\affiliation{%
  \institution{University of Waterloo}
  \city{Waterloo}
  \country{Canada}}
\email{edith.law@uwaterloo.ca}

\renewcommand{\shortauthors}{C. Blair, K. Larson, E. Law}

\newcommand{\gentxt}[1]{{\color{darkgray}\small\fontfamily{}\selectfont
#1}}

\newcommand{\crdraft}[1]{{\color{black}{#1}}}

% \begin{abstract}
% The diversity of human values and preferences and their personal nature present a significant challenge when aligning artificially intelligent (AI) agents, as different people often have conflicting views about how an agent should act. Commonly used methods for alignment typically involve gathering and aggregating feedback from a large group of people, a strategy that often neglects minority preferences. We argue that in some cases, agents should be aligned to individuals rather than groups, for example, if an agent is in a personal home. Accordingly, we present our solution for aligning agents to individual preferences and values, \systemname{}. Our approach involves an accessible and user-friendly chat system that allows users to specify their personal desired behaviours for AI agents. To better understand each user's unique preferences, we utilize active learning in response to videos and discussions designed to engage users in reflection about their values and preferences. We evaluate our system through a study involving 21 individuals, focusing on their personal definitions of ``respectful'' behaviour. We found that different people have vastly different definitions of respectful behaviour and that our system is able to accurately capture each individual's personal definition of respectful behaviour.
% \end{abstract}

\begin{abstract}

Aligning AI agents with human values is challenging due to diverse and subjective notions of values. Standard alignment methods often aggregate crowd feedback, which can result in the suppression of unique or minority preferences. We introduce Interactive-Reflective Dialogue Alignment, a method that iteratively engages users in reflecting on and specifying their subjective value definitions. 
This system learns individual value definitions through language-model-based preference elicitation and constructs personalized reward models that can be used to align AI behaviour. 
We evaluated our system through two studies with 30 participants, one focusing on ``respect'' and the other on ethical decision-making in autonomous vehicles. Our findings demonstrate diverse definitions of value-aligned behaviour and show that our system can accurately capture each person's unique understanding. This approach enables personalized alignment and can inform more representative and interpretable collective alignment strategies.
\end{abstract}




\begin{teaserfigure}
  \includegraphics[width=\textwidth]{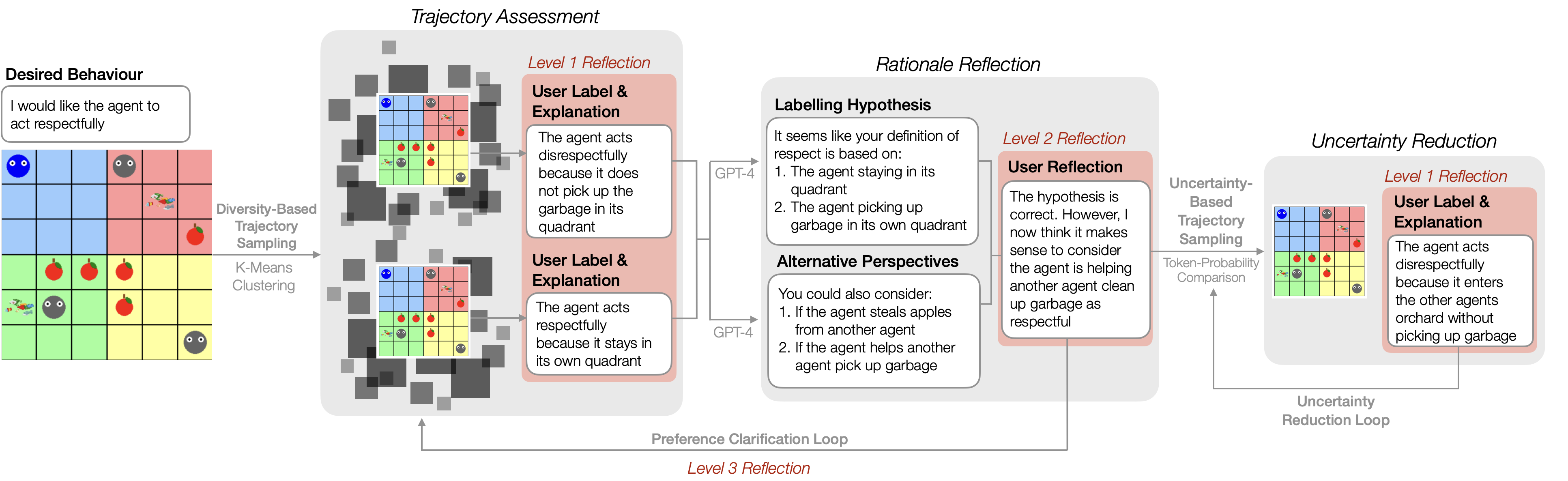}
  \caption{Dialogue-Based Preference Elicitation Portion of our Reward Design System.}
  \Description{This diagram presents a model for user preference elicitation comprising three connected sections: Trajectory assessment, rationale reflection and uncertainty reduction. On the left, the Desired Behaviour box indicates a user's intent for an agent to act respectfully. To the right of this are labelled processes within the trajectory assessment, which include steps like 'User Label \& Explanation' in response to trajectories of an agent, and 'Uncertainty Reduction Loop'. Central to the diagram, the Feature-Based Preference Elicitation Loop involves 'Labelling Hypothesis', 'User Reflection', and 'Alternative Perspectives' with nested bullet points detailing specific actions like "The agent didn't pick up the passenger in a snow jacket." The diagram's utilizes boxes with arrows mainly moving to the right, showing the flow of processes and feedback loops, and includes a preference clarification loop as well.
}
  \label{fig:teaser}
\end{teaserfigure}

\maketitle

\section{Introduction}
%\carter{Emphasize the relationship between alignment and reward modelling?}
As AI agents take on more tasks and affect personal parts of our lives, it is increasingly important to align their behaviour with human values at both individual and collective levels. This alignment challenge spans from personalized interactions, like an AI agent's respectful behaviour in a household, to broader societal collective decisions.

Current AI alignment approaches, such as reinforcement learning from human feedback (RLHF), often rely on feedback aggregated from many users \cite{casper_open_nodate, bai_training_2022}. This aggregation implicitly makes collective decisions about AI behaviour, potentially marginalizing minority viewpoints or unique preferences \cite{siththaranjan2023distributional}. By ``averaging out'' personal differences, these methods risk creating agents that ignore the values of minority groups or those with uncommon preferences. Accurate individual reward models offer a solution to this problem. These models enable personalization by tailoring AI behaviour to each user's unique values and beliefs when appropriate while simultaneously providing a foundation for more representative and interpretable collective decision-making. By understanding individual stakeholder preferences, we can aggregate diverse viewpoints more fairly and make informed trade-offs in group contexts. This approach addresses both the need for personalized AI interactions and the challenge of making ethical collective decisions.

However, creating personalized reward models presents several challenges. First, finding the most effective way for humans to convey their goals and desires to AI systems remains an open question \cite{casper_open_nodate}. Second, end users of AI systems may lack the technical skills or necessary training to provide feedback to an agent \cite{bai_constitutional_2022}. Third, it is impractical to expect individuals to teach an agent how to behave when many examples may be required.

In this work, we present \systemname{}, an interactive system for aligning AI agents to individual values that is novice-friendly and sample-efficient. The user-facing side is a simple chat interface that prompts users to explain their desired behavioural patterns and selectively sends examples of the agent's behaviour to solicit feedback using active learning techniques \cite{mosqueira-rey_human---loop_2023}. The system's textual messages are designed to encourage users to reflect deeply on their value definitions, inspired by prior work on designing for reflection \cite{fleck2010reflecting, wolfbauer_script_2022, kocielnik_reflection_2018, bentvelzen_revisiting_2022}. Using the feedback gathered from the dialogue, we create a language-based reward model that leverages the in-context learning abilities of large language models \cite{brown_language_2020}.

We evaluated \systemname{} through two studies involving a total of 30 participants. In the first study, 21 participants used the system to build a reward model for their personal definitions of respectful behaviour. The second study involved 9 participants and focused on ethical decision-making in autonomous vehicles. We found that participants had widely varying definitions of value-aligned behaviour across both studies and that our system could capture these subjective and personal definitions significantly more accurately than baseline systems.

Our contributions are as follows:
\begin{itemize}[topsep=0pt]
\item A novel, accessible, and theoretically grounded pipeline for aligning AI agents to individual values and preferences, drawing upon insights from AI, HCI, and social science research.
\item A comprehensive evaluation of the system across two distinct domains, demonstrating its ability to capture individual human values and ethical preferences.
\item Empirical evidence highlighting the diversity of individual interpretations of value-aligned behaviour.
\item Insights for future work on enabling end users to interactively align AI agents with their personal values, including potential applications in both individual and collective alignment contexts.
\end{itemize}
\section{Background and Related Work}

Our work intersects human-computer interaction, reinforcement learning, and value alignment. We draw on insights from AI, HCI, and social science research to develop a system that accurately captures individual values in a reward model. Below, we outline the related work most relevant to our system and provide the necessary background to understand it.

\subsection{Human Values and Fuzzy Preferences}
Values are principles that guide human behaviour and ethical judgments \cite{schwartz1992universals}. Friedman et al. describe values as ``what a person or group of people consider important in life'' \cite{friedman2013value}. In the context of AI systems, values can manifest in various ways. For example, in a household setting, the value of respect might manifest as an AI assistant using appropriate language or maintaining privacy. In the context of autonomous vehicles, values might include prioritizing passenger safety while also considering the welfare of pedestrians and other road users. Importantly, values vary across cultures, individuals, and contexts \cite{le2009values}. 

While values themselves may be clearly stated, their interpretation and application often become fuzzy when considered in specific contexts \cite{van2013translating}. This ambiguity is further compounded when attempting to translate these human values into computational terms for AI alignment \cite{russell2019human}. Users may struggle to articulate their values or determine which AI behaviours align with their values.

The concept of fuzzy preferences \cite{ribeiro1996fuzzy} provides insight into this challenge. Fuzzy decision-making acknowledges that preferences, attributes, and objectives in decision processes can be imprecise. In AI alignment, this fuzziness manifests in the difficulty of precisely defining and measuring value alignment. Users may not have clear, \textit{a priori} knowledge of what constitutes aligned behaviour, and the importance of different behavioural features may be unclear or change based on context.

Our work addresses these challenges by developing a system that helps users articulate and refine their fuzzy value definitions, enabling more precise alignment of AI systems with individual human values.

\subsection{Designing for Reflection}
Reflection has several notable benefits that can help users clarify their understanding of their preferences and values. For example, it has been found that prompting medical practitioners to reflect can increase diagnostic accuracy \cite{costa2019effects, fernandes2021adding, mamede2008effects}. Moreover, in consumer research, it has been found that prompting consumers to engage in preference self-reflection can lead to more accurate reporting of preferences and that engaging consumers in realistic decisions can increase preference elicitation accuracy \cite{hauser2014self}.

Designing for reflection has been of interest to HCI researchers for some time \cite{price2003new, yukawa2003co, rogers2006framework, lindstrom2006affective, ghajargar2018designing}. Fleck \& Fitzpatrick propose a framework for designing for reflection that includes definitions of various levels of reflection and techniques for supporting reflection \cite{fleck2010reflecting}. They define five levels of reflection, ranging from simply revisiting past events to critical reflection that considers wider social or moral contexts. Levels 1-3 are most relevant to our work as they inspired the design of our system: Level 1 involves explaining or justifying actions/events. Level 2 involves exploring relationships and considering different perspectives and hypotheses. Level 3 involves fundamentally challenging assumptions and transforming one's understanding or practice.

LLMs present an opportunity for more tailored and dynamic dialogue to engage users in reflection. Some work has explored the use of LLMs for generating reflective prompts in a design template \cite{yin_jamplate_2024}, while others have used language prompts to engage users in reflection through single questions or sentence starters \cite{fessl2017known, renner2016effects, williams2016revising, ifenthaler2012determining}. Further extensions of this approach have included the use of pre-scripted, multi-message dialogues to facilitate reflection~\cite{kocielnik2018reflection, wolfbauer2022script}.

We extend this body of work by integrating reflective dialogue techniques with large language models, creating a dynamic and personalized reflection process that helps users clarify their values for AI alignment.

\subsection{Reinforcement Learning and Its Variants}
Reinforcement Learning (RL) is a branch of machine learning in which an agent learns a policy, i.e., a strategy for selecting actions in an environment, to maximize the rewards it receives. Traditionally, the agent designer crafts a reward function to incentivize desirable behaviour. However, reward design is notoriously challenging and can result in reward hacking where an agent learns behaviour that gets high reward, but that runs counter to the designers intent \cite{amodei_concrete_2016, dewey2014reinforcement, everitt2016avoiding}.

Inverse reinforcement learning (IRL) was developed in response to difficulties with reward design. In IRL, the goal is to learn a reward function from expert demonstrations \cite{ng2000algorithms, ramachandran2007bayesian, abbeel2004apprenticeship, ziebart2008maximum, neu2009training}. This can be useful when a desired behaviour is easy to demonstrate but difficult to design a reward function for. However, since multiple reward functions can often explain the observed behaviour, the reward function lacks interpretability and is inherently ambiguous \cite{metelli2023towards}.

Reinforcement learning from human feedback (RLHF) instead learns a reward model from human feedback \cite{christiano_deep_2017, ziegler2019fine, bai_training_2022}. This approach is especially useful when aligning agents to human preferences as the agent can optimize for getting ``good'' feedback. There are three main parts of RLHF: Feedback collection where humans provide feedback, reward modelling where the feedback is turned into a reward model, and policy optimization where the reward model is used to train a policy/agent with reinforcement learning \cite{casper_open_nodate}.

Our approach bridges the gap between RL and HCI by combining RLHF techniques with user-friendly interfaces and reflective dialogue, making the process of aligning AI agents with individual values more accessible to non-expert users.

\subsection{Preference Elicitation and Active Learning}
When using language feedback for RLHF, it is natural to implement dialogue-based preference elicitation as the mechanism for collecting feedback. In dialogue-based preference elicitation, the goal is to understand a user's preferences through the use of dialogue \cite{priyogi_preference_2019, radlinski_coached_2019}. There are two main approaches to preference elicitation: item-based preference elicitation, where the user is asked about their preferences regarding specific things \cite{adomavicius2005toward, sarwar2001item, loepp2014choice, christakopoulou2016towards, graus2015improving} and feature-based preference elicitation where the user is asked about general features or attributes \cite{llorente2011increasing, sun2018conversational, viappiani2006preference}. Some methods use both item- and feature-based methods \cite{biyik2023preference}, similar to our approach. However, previous work has focused on content recommendation, not agent alignment.

When performing item-based preference elicitation, we must select items to query the user about. Active learning is a subfield of machine learning where the learner is responsible for choosing which examples to request labels for \cite{settles2009active, ren2021survey}. Being selective about which examples to query the user about is important in cases where labels are expensive to obtain \cite{olsson2009literature}. There are generally three strategies for sampling items to query the user about, namely random-, diversity-, and uncertainty-based sampling \cite{monarch2021human}. Random sampling selects items arbitrarily, while diversity-based sampling chooses items that are most different from each other. Uncertainty-based sampling, in contrast, focuses on items the model is least confident about~\cite{mosqueira-rey_human---loop_2023}.

Our system integrates these approaches into an interactive dialogue framework, combining item- and feature-based preference elicitation with active learning and reflective dialogue to capture individual value definitions efficiently.

\section{Interactive-Reflective Dialogue Alignment (IRDA) System}

\begin{figure*}
  \centering
  \includegraphics[width=\textwidth]{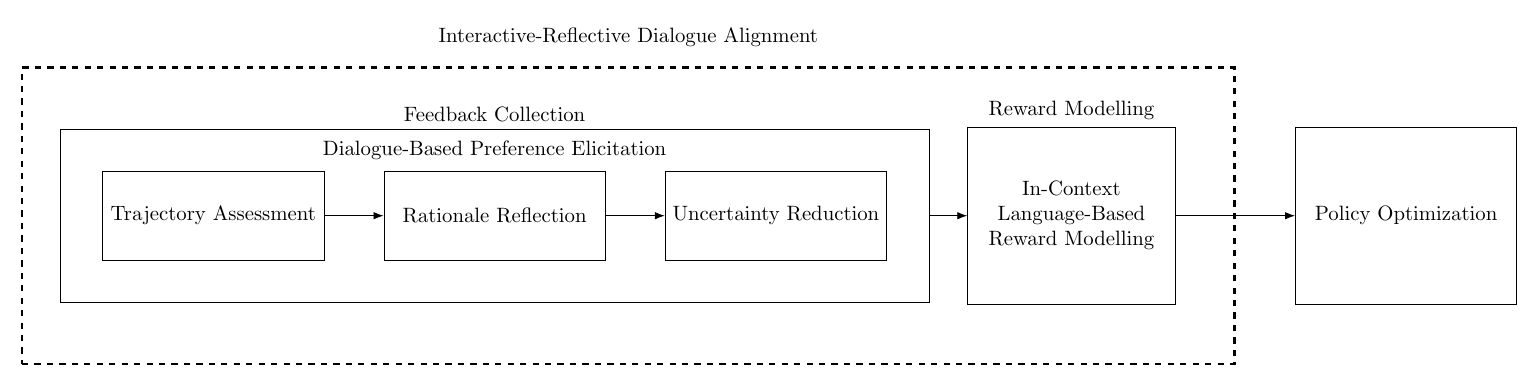}
  \caption{Overview of the RLHF pipeline with \systemname{}.}
  \Description{A flowchart showing the process from Feedback Collection to Policy Optimization. Feedback Collection includes Dialogue-Based Preference Elicitation, which consists of three sequential steps: Trajectory Assessment, Rationale Reflection, and Uncertainty Reduction. This is followed by Reward Modelling using In-Context Language-Based Reward Modelling, leading to Policy Optimization.
}
\end{figure*}

We introduce an interactive system, named \systemname{} (IRDA), which aims to enable end users with no particular expertise in machine learning to define agent behaviour that is aligned with their subjective definitions of a human value and generate a reward model that can be later used to train an agent based on this definition. We will begin by outlining our design aims, providing a high-level overview of the system and user flow, and then describe each component in more detail.

\subsection{Design Aims}
Our system is intended for the end-users of AI agents, and we do not assume that the users of our system will have any technical expertise. As such, the design aims for our system are centred around the idea of making the system easy and quick to use while maintaining good performance. In particular, we require:
\begin{enumerate}
    \item The system should not require any special technical knowledge (e.g., how to program or how to design a ``good'' reward function).
    \item The system should be sample-efficient (i.e., strategically selecting the most informative examples to ask users for feedback on).
    \item The system should be able to capture individual differences and unique conceptions of the appropriate behaviour associated with a value.
\end{enumerate}

\subsection{System and User Flow}\label{section:user_flow}

%Our IRDA system consists of three components: a user-facing \textbf{chat interface} (shown in Figure~\ref{system_screenshot}), which allows the user to interact with the system; a \textbf{value elicitation module}, which generates dialogue to engage users in a reflective conversation aimed at helping users articulate their own definition of a value, and a \textbf{reward modelling module} which uses the information gained in conversation with the user to create a reward model.  Together, these components form a system that facilitates the first two stages of Reinforcement Learning from Human Feedback (RLHF), namely feedback collection and reward modelling. The third stage of RLHF is policy optimization where the reward model is used to train an agent to adopt value-aligned behaviours.  Policy optimization is not part of our system and is outside the scope of this paper, but the reward models generated by our system are compatible with nearly all policy optimization methods.  

%The input to our system is a human value (e.g., ``respect") and a reinforcement learning environment, and the output is a reward model can train an agent to adopt behaviour that aligns with that value (e.g., act respectfully). 

Upon entering our system, users are presented with a greeting message outlining the purpose of the system as well as a preview of the environment. To better illustrate our system, we will use one of the environments used in our evaluation, the \textit{Multi-Agent Apple Farming Environment}, as the running example. The environment is a 6$\times$6 grid where the agents are rewarded for picking apples and receive no reward for collecting garbage. There is one blue ``main'' agent in the environment and three grey ``background'' agents as shown in Figure~\ref{fig:teaser}. Each one of the agents owns one of the four 3$\times$3 quadrants, which each represent an orchard. Each agent is free to move around the whole grid; however, two of the three background agents are programmed to be stationary. The main agent is the agent whose behaviour users are asked to monitor and give feedback on. Users give feedback on whether the behaviour aligns with a particular value (e.g., picking apples but being ``respectful'' to neighbours).

%The user flow and how it is supported by the system is described below (and is shown in algorithmic form in \ref{alg:IRDA} in the appendix).

\begin{figure}
  \centering
  \includegraphics[width=0.98\linewidth]{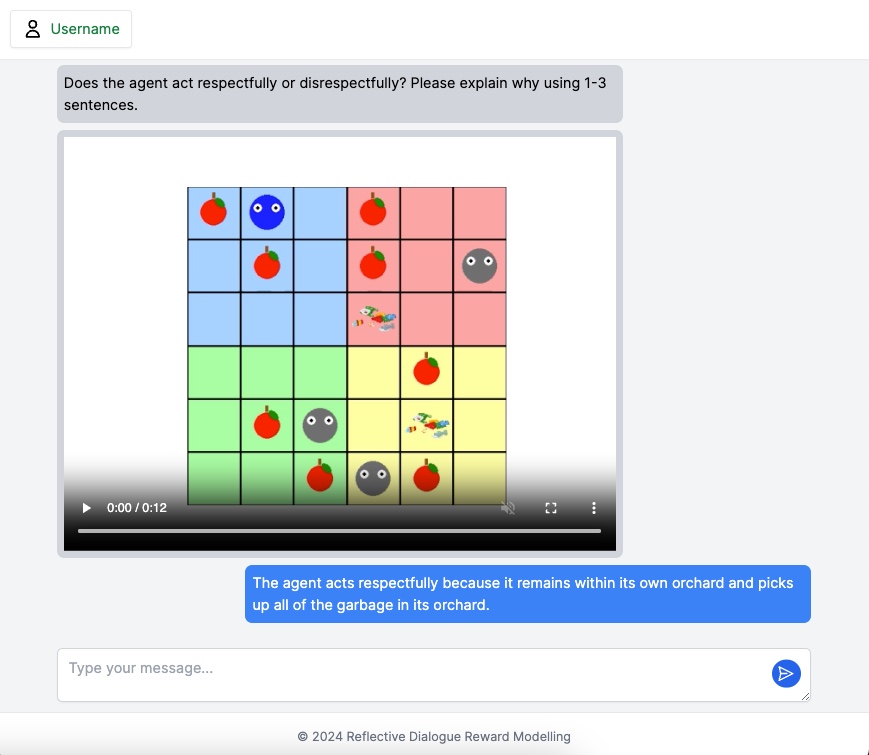}
  \caption{System screenshot showing the user-facing chat interface of the \systemname{} system.
  }
  \Description{Screenshot of a dialogue interface for evaluating agent behaviour. The image shows a grid-based game environment with apples, avatars, and trash items. The system asks about the agent's respectfulness, and a response from the user indicates the agent acts respectfully by staying in its orchard and cleaning up garbage. The interface includes a video player and a text input field for user responses.
}
  \label{system_screenshot}
\end{figure}

The input to our system is a value that the user wants the agent to adopt. %Users are then asked to briefly explain how they would like the agent to act.
% To begin, we allow the user to specify how they would like the agent to act in natural language. 
For example, a user could specify to the system ``I would like the agent to act respectfully'' through a chat interface, as shown in Figure~\ref{system_screenshot}. Given this value specification, we collect a large pool of trajectories (sequences of actions the agent took in the environment). This pool can be collected by taking a portion of an existing dataset or by running random rollouts in a simulator. The system then selectively samples a small number of trajectories where the agent behaves in diverse ways to present to the user for feedback ({\bf trajectory assessment}). Based on this feedback, the system hypothesizes about the rationale behind the user's assessments (e.g., the user thinks the agent's behaviour is disrespectful because it wandered into another agent's yard). This hypothesis is presented to the user along with alternative perspectives the user could consider to prompt them to reflect on their current value definition ({\bf rationale reflection}).  Upon reflection, the user can opt to re-explain the trajectories they initially saw with their new perspective. This iterative process is called the preference clarification loop (shown in \autoref{fig:teaser}).

All of the information collected from the user through this iterative feedback loop is then used to build an initial language-based reward model. The initial reward model is refined by selectively picking trajectories that the model is most uncertain about and querying the user for feedback ({\bf uncertainty reduction}). Each time the user explains one of the trajectories the model is uncertain about, this information is added to the reward model, thus making it less uncertain about similar trajectories. This process is repeated until the overall uncertainty is below a certain threshold.  In the end, the system generates a language-based reward model that can effectively administer rewards to any agent when it acts in accordance with the user's value definitions ({\bf reward modelling}). Depending on the complexity of the desired behaviour and environment, this reward model can then be used to either train a more efficient reward model or to train an agent directly.

Having described, at a high level, the system and user flow, we will now describe various user flow steps (those in bold) in more detail, namely trajectory assessment, rationale reflection, uncertainty reduction and reward modelling.\\

%\subsection{Feedback Collection} %\label{feedback_collection}

%\noindent {\bf Feedback Collection.} Our system works with the user, through a natural language chat interface (Figure~\ref{system_screenshot}), to refine their value definition and understand agent behaviours that align with that definition.  This refinement process is facilitated through two types of conversational interaction between the system and the user: {\bf trajectory assessment} (where the user are presented with a behavioural trajectory and asked to assess whether it conforms to their value definition) and {\bf rationale reflection} (where users are asked whether certain agent actions or environment features can explain their trajectory assessment).  

% \noindent {\bf Trajectory Assessment.}
\subsubsection{Trajectory Assessment}\label{section:trajectory_assessment}
\begin{figure}
  \centering
  \includegraphics[width=0.98\linewidth]{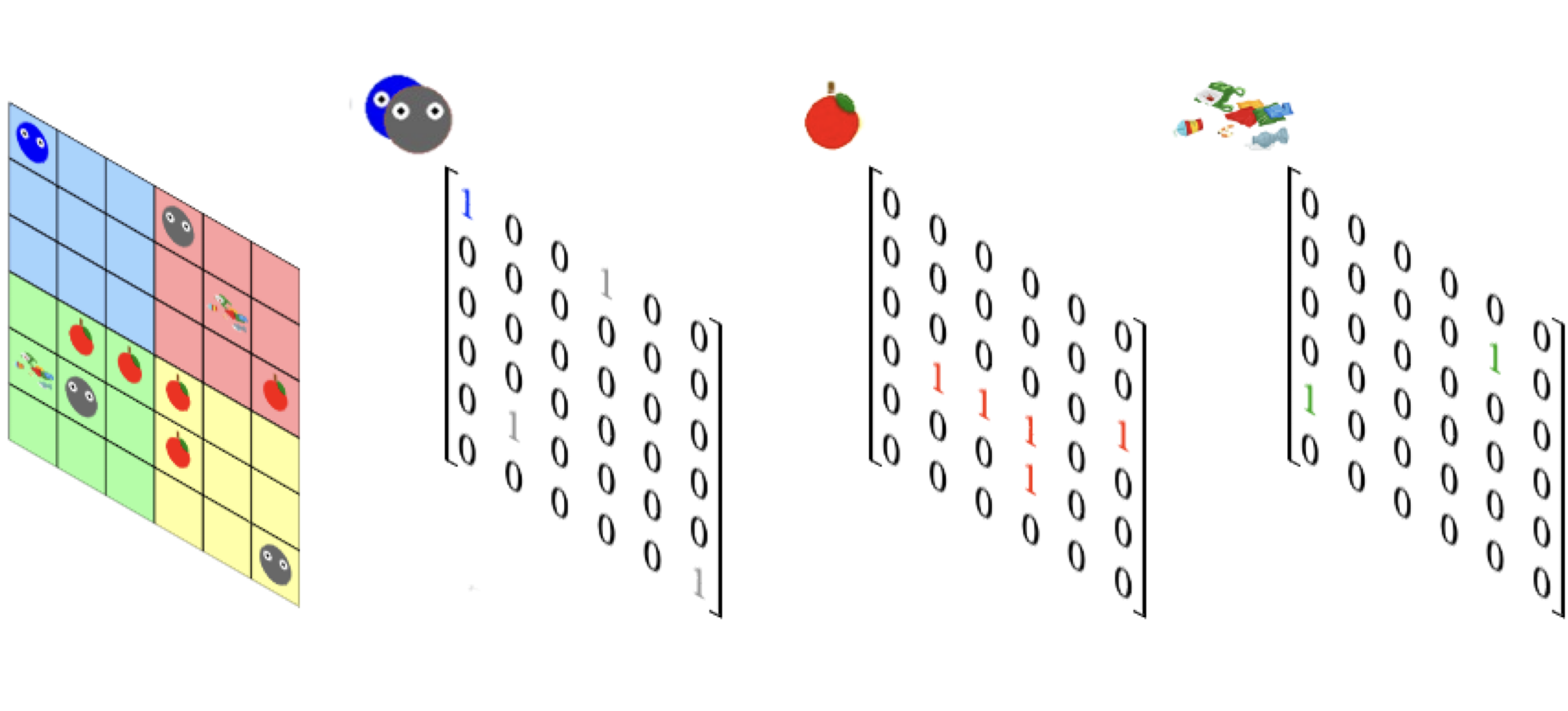}
  \caption{A numerical encoding of one timestep of the multi-agent apple farming environment. Agent positions are encoded in one array, apple positions in a second array, and garbage positions in a third.
  }
  \Description{The image shows four panels side by side. The leftmost panel displays a grid-based game environment with colored sections, apples, avatars, and trash. The three rightmost panels represent binary matrices, each associated with a different game element (avatar, apple, and trash). }
  \label{numerical_encoding}
\end{figure}

Our system initially lacks information about the user's preferences, and thus, we employ \textit{diversity-based sampling} \cite{nguyen2004active} to find a set of trajectories where the agent exhibits a diverse set of behaviours in varying situations. To do so, we begin by sampling a large pool of over 1000 trajectories. We turn each trajectory into a numerical array that gives a full representation of the state of the environment at each time step. For example, in a 30-step trajectory in the multi-agent apple farming environment, we can encode each step as multiple arrays of the same shape as the grid. In each array, for a step, we can encode the position of a certain type of entity within the environment. The numerical encoding for a single timestep is shown in Figure~\ref{numerical_encoding}.

With the pool of numerically encoded trajectories in hand, we perform $k$-means clustering to split the trajectories into $k$ distinct clusters. Since we want to query the user about a diverse set of trajectories in this initial phase, we select one trajectory from each of the $k$ clusters. In particular, for each group, we select the trajectory closest to the arithmetic mean, or the centroid, of all the trajectories in the cluster.

The system implements item-based preference elicitation by asking the user to explain whether the agent is aligned with their preferences in each of the $k$ trajectories, one at a time, using 1-3 sentences. For instance, if a user were trying to train an agent to behave respectfully, the system would ask, ``\texttt{Does the agent act respectfully? Please explain your reasoning using 1-3 sentences.}''\\

%In the spirit of making our system accessible to users with varying backgrounds and skill sets, we collect feedback from users in natural language through a chat interface shown in Figure~\ref{system_screenshot}.
% In addition to being a common and accessible mode of interaction, 
%Dialogue is an engaging mode of interaction~\cite{ruan2019quizbot, wolfbauer_script_2022}, making it an appealing feedback collection method. We frame our language-based feedback collection as \textit{dialogue-based preference elicitation}, where we collect feedback about users' preferences regarding the agent's behaviour. The dialogue is generated by the \textbf{value elicitation module}.

%\subsubsection{Dialogue-Based Preference Elicitation}\label{dialogue-based_preference_elicitation}
%We employ both item-based and feature-based preference elicitation in multiple stages outlined below. The dialogue-based preference elicitation process is shown in Figure \ref{fig:teaser}. 

%\subsubsection*{Stage 1: Initial Item-Based Preference Elicitation.} To perform item-based preference elicitation, we first select items (specific examples of agent behaviour) to query the user about. 

%\subsubsection*{Stage 2: Feature-Based Preference Elicitation.}

% \carter{Possibly explain trajectory encoding here?}
% \noindent {\bf Rationale Reflection.}
\subsubsection{Rationale Reflection}
\begin{figure}[htbp]
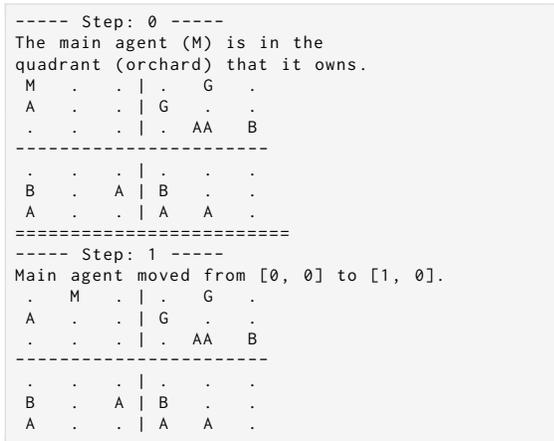

    \centering
    \begin{minipage}{0.4\textwidth}
    \begin{lstlisting}
----- Step: 0 -----
The main agent (M) is in the 
quadrant (orchard) that it owns.
 M   .   . | .   G   .
 A   .   . | G   .   .
 .   .   . | .  AA   B
-----------------------
 .   .   . | .   .   .
 B   .   A | B   .   .
 A   .   . | A   A   .
=========================
----- Step: 1 -----
Main agent moved from [0, 0] to [1, 0]. 
 .   M   . | .   G   .
 A   .   . | G   .   .
 .   .   . | .  AA   B
-----------------------
 .   .   . | .   .   .
 B   .   A | B   .   .
 A   .   . | A   A   .
    \end{lstlisting}
    \end{minipage}
    \caption{ASCII encoding of two timesteps of a trajectory of the multi-agent apple farming environment.}
    \Description{ASCII representation of agent movement in a grid environment. The image shows two steps (Step 0 and Step 1) of a scenario where the main agent (M) moves within its quadrant. The grid is divided into sections, with various elements represented by letters: M for main agent, G for garbage, A for apples, and B for other agents. The figure illustrates the agent's initial position and its movement from [0, 0] to [1, 0].
}
    \label{fig:ascii-encoding}
\end{figure}

During the trajectory assessment process, the system intermittently interjects a dialogue to engage the users to reflect on their value definition. To do this, we use a large language model (GPT-4 Turbo) to make a hypothesis about what features the user is basing their decisions on and other features the user could consider. To do so, we parse each trajectory into an ASCII representation (see Figure \ref{fig:ascii-encoding} for an example) and pass this along with the user's explanation to the LLM. In the prompt, we ask the LLM to make a hypothesis about what features the user is basing their decisions on and to offer alternative features the user could consider.

For example, in the multi-agent apple farming environment, where an agent owns one of four orchards in the environment, the user may base their definition of ``respectful'' behaviour on whether the agent stays in its own orchard. A simplified example of a response from the system could be:
\begin{quote}
    \texttt{Based on your explanations, it seems as though a key factor in determining whether the agent's behaviour is respectful or not is whether the agent stays in its own orchard.}
    \texttt{You could also consider:}
    \texttt{
    \begin{enumerate}
        \item Whether the agent helps pick up garbage
        \item Whether the agent steals apples from other agents
    \end{enumerate}
    }
\end{quote}

The user is then prompted to explain if the hypothesis the system made is correct and if the other features should be considered. This feature-based preference elicitation step is important for a number of reasons. First, understanding the features on which the user bases their decisions allows for improved generalization of the reward model. The system can use these features, or patterns of behaviour, as decision criteria when assessing whether the agent's behaviour in a trajectory aligns with the user's value definition. Second, this step encourages the user to reflect on their definition and consider alternative perspectives. This reflection can assist users in gaining a clearer understanding of the specific behaviour they want the agent to demonstrate. Third, if the diversity-based sampling in the initial item-based preference elicitation phase does not select any trajectories where a certain behavioural feature is present that would be important to the user, the alternative or additional features proposed by the system can help uncover these. This can help the system gain a more holistic understanding of the user's preferences, even when only querying the user about a handful of items.

If the user updates the features they think are important based on the system's proposed alternatives, they can optionally re-assess the initial $k$ trajectories in the preference clarification loop. This allows users to iteratively reflect on and specify  what features or behavioural patterns they would like the agent to embody. This refinement process simultaneously helps the user better understand their own preferences and helps the system gain a clearer understanding of the user's true intent.\\

%\subsubsection*{Stage 3: Uncertainty-Reduction Item-Based Preference Elicitation.}

% \noindent {\bf Uncertainty Reduction.}
\subsubsection{Uncertainty Reduction}
% Once the user has provided feedback alternating between trajectory assessment and rationale reflection, the system creates an \textit{initial} reward model based on the user's feedback so far. The reward model can be thought of as a classifier - the input is an ASCII representation of a trajectory, and the output should be a 1 if the agent's behaviour aligns with the user's expressed intent and 0 otherwise. 

After the user has provided feedback through the alternating processes of trajectory assessment and rationale reflection, we implement another phase of item-based preference elicitation to create and refine the reward model. This approach focuses on querying the user about specific items (trajectories) to improve the model's understanding of user preferences. 

To begin with, we use all of the information collected in the trajectory assessment and rationale reflection phases to create an initial reward model. The reward model can be thought of as a classifier - the input is an ASCII representation of a trajectory, and the output should be a 1 (reward) if the agent's behaviour aligns with the user's expressed intent and 0 (no reward) otherwise. We use this initial model to assign rewards to a held out pool of trajectories. Our reward model, based on an autoregressive transformer (LLM), allows us to identify trajectories where the model has high uncertainty about the appropriate reward. We do this by comparing token probabilities for positive (reward) and negative (no reward) classifications. For example, if the user is training the agent to be respectful, we compare the probability of the reward model outputting the "respectful" token to the probability of the model outputting the "disrespectful" token. If the probabilities are close to one another, then the model is less certain, and if they are very different (e.g. 0.99 and 0.01), then the model has high certainty about whether the agent acts according to the user's intent. The "confidence" of the model can be thought of as the absolute value of the difference between the two token probabilities.

The item-based preference elicitation process for refining the reward model proceeds as follows: (1) Identify the trajectory in the pool where the model has the lowest confidence. (2) Query the user to explain this trajectory. (3) Add the ASCII representation of the trajectory and the user's explanation to the reward model. (4) Repeat steps 1-3 until the model's confidence for each trajectory in the uncertainty pool exceeds a minimum confidence threshold, $\epsilon$.

This iterative process reduces the model's uncertainty about the user's value definition by focusing on specific items (trajectories) where additional user input is most beneficial.
\subsubsection{Reward Modelling}
Once the feedback from the user has been collected, the last step is to create a reward model that can give feedback to the learning agent on behalf of the human. Creating a reward model that can issue rewards on behalf of humans is important as, depending on the complexity of the environment and the desired behaviour, it can take millions~\cite{pardo2018time} or even billions~\cite{agapiou2022melting} of timesteps for an agent to learn the desired behaviour. The reward model can act on behalf of the user, thus relieving the user of giving feedback during agent training.
 
%\subsubsection{In-Context Language-Based Reward Model}

% The reward model can be thought of as a classifier in that it receives a trajectory parsed into ASCII (see \autoref{fig:ascii-encoding}) as input and outputs $1$ if the agent's behaviour aligns with the user's expressed intent and $0$ otherwise. To achieve this, we provide an LLM with the information collected about the user during the dialogue-based preference elicitation and ask the LLM whether the agent's behaviour in the input trajectory is aligned with the user's expressed intent. If the LLM thinks the agent behaves according to the user's expressed intent, we parse this to a 1 and otherwise a 0.

Our in-context, language-based reward model functions as a classifier, evaluating agent behaviour based on user-defined values. The model receives as input a trajectory encoded in ASCII format, a process detailed in \autoref{appendix:trajectory_encoding}. This encoding preserves essential spatial and temporal information while enabling efficient processing by language models (see \autoref{fig:ascii-encoding} for an example).
The model leverages two key elements to classify a trajectory: (1) The encoded trajectory and (2) the user feedback collected during dialogue-based preference elicitation.

Using this information, we prompt the LLM to assess whether the agent's behaviour in the given encoded trajectory aligns with the user's expressed intent. The LLM's output is then parsed into a binary classification: 1 if the behaviour aligns with user intent, 0 otherwise.

To achieve this, we create a prompt that includes 
\begin{enumerate}
    \item a description of the environment and the ASCII characters used in the encoding,
    \item the information gained from the user during the dialogue-based preference elicitation,
    \item a description informing the language model that its goal is to predict whether the agent's behaviour is aligned with the user's value definition,
    \item an ASCII representation of the trajectory that the model is to label,
    \item text to encourage the model to engage in chain-of-thought reasoning,
    \item and instructions about how the LLM is to format its response so that the output can be programmatically parsed.
\end{enumerate}

Two major findings about LLM capabilities inspired the design of this prompt. First, LLMs are effective in-context, few-shot learners \cite{brown_language_2020}. This means that a trained LLM, with fixed parameters, can learn new patterns by including the beginning of a pattern as context and asking the model to continue the pattern. In settings similar to ours, this has been shown to be far more sample efficient than traditional supervised learning~\cite{kwon2022reward}. The pattern of interest, in our case, is what behaviour the user deems to be aligned with their value definition.
The second finding is that prompting LLMs to engage in chain-of-thought reasoning can greatly increase their performance~\cite{kojima2022large}. The idea is that the model forms an argument first and then, due to the auto-regressive nature of language models, uses the argument it made to determine a final answer. 
% In other words, the LLM must justify its answer before giving it.

Since our LLM-based reward model receives a full trajectory as input and outputs a reward, it can deal with non-Markovian rewards. This means that the reward model can evaluate the agent's behaviour over multiple time steps when determining the reward. We hypothesized that this would be important for capturing users' preferred agent behaviour, and this was supported by our empirical findings discussed in Section \ref{section:beyond_markov}.
% For example, some users thought that it was only respectful for an agent to enter another agent's territory if that other agent had already entered their territory. This is markedly non-Markovian as it cannot be determined in a single timestep.

\section{Study Design \& Methodology}
% We evaluated our system, \systemname{} by comparing it to two baselines in two different environments: a multi-agent apple farming environment and the moral machine setting \cite{awad2018moral}. 
The goal of our system, \systemname{}, is to learn individual value definitions. We evaluated our system in two studies by comparing to another language-based reward modelling pipeline and to supervised learning. Study 1 investigates the utility of our system for learning about participants' definition of \textit{respectful} agent behaviour. Study 2 investigates the utility of our system for learning about participants' decision-making in moral dilemmas involving an agent (autonomous vehicle). Our studies employ a within-subject design, collecting data from each participant to train various language-based reward models, supervised learning baselines, and for a test set to evaluate these methods. 
% This section describes the methodology common to both studies, with specifics provided in their respective sections.

\subsection{Environments}
In Study 1, the multi-agent apple farming grid world, which was described in the beginning of \autoref{section:user_flow}, was used. Participants were asked to evaluate if the agent was acting respectfully.

In Study 2, we used the Moral Machine environment \cite{awad2018moral}. The Moral Machine dataset is based on a simulated environment that presents ethical dilemmas faced by autonomous vehicles. In each scenario, a self-driving car encounters an unavoidable accident and must choose between two outcomes: staying on course or swerving. Each decision results in different consequences for the individuals involved. The environment populates these scenarios with characters chosen from 20 predefined types, including pedestrians, passengers, and various demographic groups such as children, adults, elderly, and animals. Each outcome features between one and five characters. The scenarios explore nine key ethical dimensions: the nature of the vehicle's intervention, the relationship of individuals to the vehicle (pedestrians or passengers), the legality of pedestrians' actions, and various attributes of potential victims, including age, gender, social status, physical fitness, number, and species (human or pet). Participants were asked to choose whether the vehicle should stay or swerve in each scenario presented to them. 

\subsection{Participants}
\begin{figure}[h]
  \centering
  \includegraphics[width=0.98\linewidth]{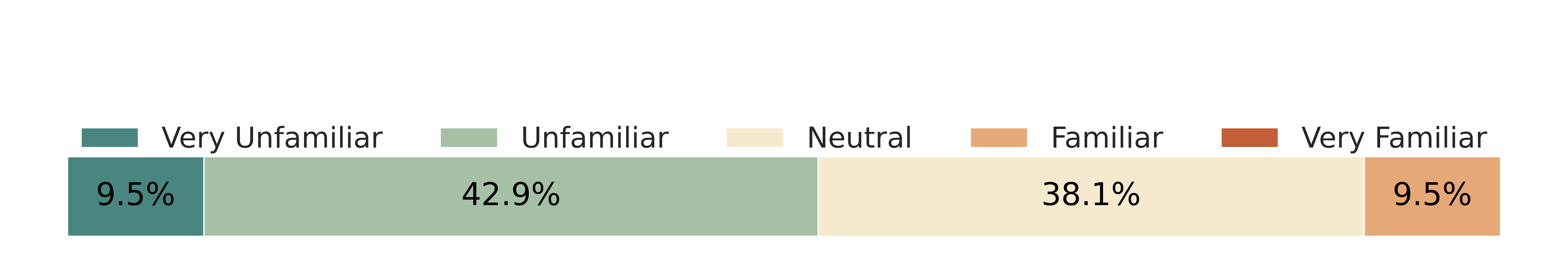}
  \caption{Participant familiarity with reinforcement learning in Study 1.}
  \Description{Horizontal stacked bar chart showing participant familiarity with reinforcement learning in Study 1. The chart displays five categories of familiarity: Very Unfamiliar (9.5\%), Unfamiliar (42.9\%), Neutral (38.1\%), Familiar (9.5\%), and Very Familiar (0\%).
}
  \label{fig:rl_familiar_s1}
\end{figure}

In Study 1, we recruited 21 participants from our institution (18 to 39 age range, M=23.86, 7 self-identified as male and 14 as female). When asked to rate their level of familiarity with reinforcement learning on a 5-point Likert ranging from ``very unfamiliar'' (1) to ``very familiar'' (5), the mean level of familiarity was 2.48, with the mode and median being 2. The Likert-scale data is visualized in figure \ref{fig:rl_familiar_s1} and highlights that more than half of participants were ``unfamiliar'' or ``very unfamiliar'' with reinforcement learning.

\begin{figure}[h]
  \centering
  \includegraphics[width=0.98\linewidth]{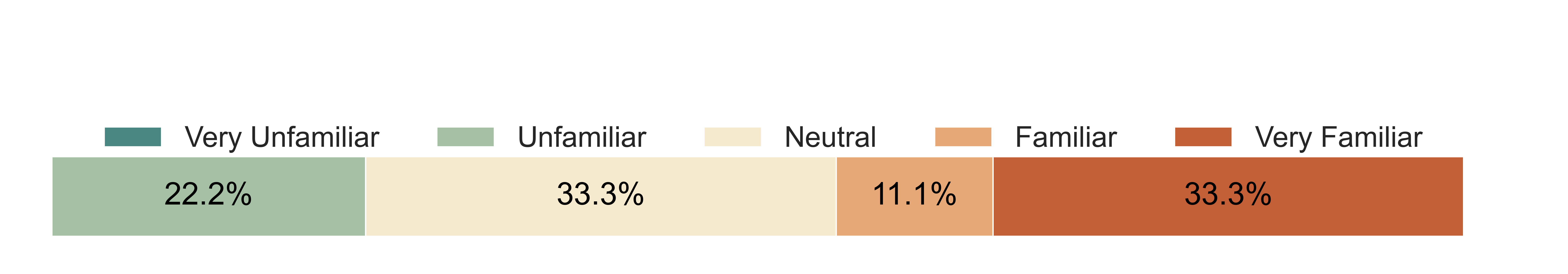}
  \caption{Participant familiarity with reinforcement learning in Study 2.}
  \Description{ Horizontal stacked bar chart illustrating participant familiarity with reinforcement learning in Study 2. The chart presents four levels of familiarity: Unfamiliar (22.2\%), Neutral (33.3\%), Familiar (11.1\%), and Very Familiar (33.3\%). }
  \label{fig:rl_familiar_s2}
\end{figure}

In Study 2, we recruited 9 participants from our institution (18 to 33 age range, M=25.66, 6 self-identified as male and 3 as female). When asked to rate their level of familiarity with reinforcement learning on a 5-point Likert ranging from ``very unfamiliar'' (1) to ``very familiar'' (5), the mean level of familiarity was 3.55, with the mode and median being 3. The Likert-scale data is visualized in figure \ref{fig:rl_familiar_s2}.

\subsection{Procedure} 
Participants used our interface to specify how they would like the agent to act via our dialogue-based preference elicitation process. Participants then labelled 50 scenarios and were interviewed.

\textit{Introduction ($\sim$5min)} - After completing the consent form and the demographic questions, the mechanics of the environment were fully explained to the user. We explained the environment mechanics as thoroughly as possible so that differences observed between participants stemmed from their opinions, not hidden assumptions about the environment.

% Following the explanation of the environment, participants were told that they would only be evaluating the blue agent that starts in the upper left quadrant and that they should assess whether this agent acts respectfully according to their own understanding.

\textit{Dialogue} - The participant began by conversing with the system about the agent's behaviour following the dialogue structure presented in Section \ref{section:user_flow}. To control the amount of time users spent on the activity, we limited the user to one preference clarification loop and one uncertainty reduction loop. 

\textit{Labelling} - Following the participants' dialogue interaction with the system, participants labelled 50 scenarios. Each participant labelled the same scenarios, which allowed us to assess how much the participants agreed on the labels.

\textit{Semi-structured Interview ($\sim$10min)} - After completing the labelling task, each participant was asked whether they felt they were able to give the system a good understanding of their decision-making if they were always able to articulate why they chose a label, if it was ever difficult to decide, and if they thought their labelling behaviour changed over time. The interview aimed to determine two things. First, it sought to determine if participants could verbally explain their definition of aligned behaviour. Second, it attempted to discover whether these definitions changed over time and what factors caused any changes.

\subsection{Baseline Comparisons}

We compared our system, \systemname{}, to several baselines to evaluate its effectiveness. These baselines include another language-based system and various supervised learning approaches.

\subsubsection{Language-Based Baseline ($L^B$)}\label{baseline}
Kwon et al. \cite{kwon2022reward} proposed a reward modelling pipeline for text-based environments where the user selects multiple examples from a palette of examples of the agent behaving as they would desire, accompanied by explanations. We slightly modify their pipeline in the following way: Instead of asking the user to select examples from a handcrafted palette, we choose the examples the user sees with the diversity-based sampling procedure described in Stage 1 of Section~\ref{section:trajectory_assessment}.

Since the baseline pipeline $L^B$ is a subprocess of our full system, \systemname{}, the user only interacts with our system. However, when forming the reward model for $L^B$, we only include the information collected from the user during the trajectory assessment phase (described in Section \ref{section:trajectory_assessment}) before they have done rationale reflection, mirroring the pipeline proposed in \cite{kwon2022reward}.

\subsubsection{Supervised Learning Baselines}
We compared our approach to supervised learning methods using neural networks. These baselines include both individual models trained separately for each participant and collective models trained on aggregated data from all participants. For more details on the architecture and training of these models, see \autoref{appendix:SL}.

In Study 1, we employed multi-layer perceptron (MLP) models: individual models ($\text{MLP}^{\text{ind}}_i$) for each participant $i$, and a collective model ($\text{MLP}^{\text{col}}$) using aggregated data from all participants. These models used input based on the grid map encoding (see \autoref{numerical_encoding}).

For Study 2, we compared to both MLP and convolutional neural network (CNN) models. We used the same MLP architecture as in Study 1, but with a 26-dimension vector input representing Moral Machine scenarios described in \autoref{appendix:trajectory_encoding}. We also introduced CNN models, both individual ($\text{CNN}^{\text{ind}}_i$) and collective ($\text{CNN}^{\text{col}}$), which use image representations of scenarios as input.

Across all supervised learning models, we utilized 30 scenarios per participant for training, selected from the 50 scenarios annotated during the labelling phase.
By comparing \systemname{} to these baselines ($L^B$, $\text{MLP}^{\text{ind}}_i$, $\text{MLP}^{\text{col}}$, $\text{CNN}^{\text{ind}}_i$, and $\text{CNN}^{\text{col}}$), we aim to evaluate the efficacy of our language-based reward modeling approach and the value added by our reflective dialogue process.

\subsection{Analysis}

Our analysis aims to answer the following three questions:
\begin{enumerate}[label=\textbf{RQ\arabic*:}]
\item Do value definitions significantly vary between participants?
\item Does structured reflection enhance language-based reward modelling?
\item When is individualized language-based reward modelling effective?
\end{enumerate}

\begin{table*}[!htbp]
\centering
\caption{Mapping of Analysis Methods to Research Questions}
\label{tab:analysis-rq-mapping}
\resizebox{0.7\textwidth}{!}{%
\begin{tabular}{>{\raggedright\arraybackslash}p{0.5\textwidth}ccc}
\toprule
\textbf{Analysis Method ($\S$)} & \textbf{RQ1} & \textbf{RQ2} & \textbf{RQ3} \\
\midrule
\rowcolor{lightgray}
Inter-Annotator Agreement (\ref{section:inter_ann_agree}) & $\checkmark$ &  & $\checkmark$ \\
Evaluation of Language-Based Reward Model Performance (\ref{section:eval_lbrm}) &  & $\checkmark$ & \\
\rowcolor{lightgray}
Comparison to Supervised Learning (\ref{section:comp_sl}) & $\checkmark$ &  & $\checkmark$ \\
Qualitative Analysis of Participant Decision Making (\ref{section:qual_decision}) & $\checkmark$ &  &  \\
\rowcolor{lightgray}
Analysis of Feature Similarity Between Participants (\ref{section:feature_similarity})& $\checkmark$ &  & $\checkmark$ \\
Thematic Analysis of Interview Data (\ref{section:thematic})&  & $\checkmark$ & $\checkmark$ \\
\bottomrule
\end{tabular}%
}
\end{table*}

To address these questions, we employ a mixed-methods approach, combining quantitative analyses of model performance and inter-annotator agreement with qualitative analyses of participant decision-making processes and experiences. The following subsections detail our analytical methods, each designed to provide insights into one or more of our research questions.

\subsubsection{Inter-Annotator Agreement}\label{section:inter_ann_agree}
We assess the inter-annotator agreement between participants on the test set of scenarios they labelled in each study. Since each participant labelled the same test scenarios, we can use Fleiss' kappa value to quantify the inter-annotator agreement between the participants \cite{landis1977measurement}. Generally, kappa statistics below 0 indicate ``poor'' agreement and kappa statistics above 0.8 indicate ``nearly perfect'' agreement \cite{landis1977measurement}. This analysis directly addresses \textbf{RQ1} by quantifying how much participants agree when labelling examples. Low agreement suggests diverse value definitions, while high agreement indicates more uniform values across participants. It also informs \textbf{RQ3} by indicating when individualized approaches might be more beneficial than collective ones.

\subsubsection{Evaluation of Language-Based Reward Model Performance}\label{section:eval_lbrm}
Our evaluation compares our system's performance against the baseline language-based reward modelling system that does not leverage dialogic (level 3) reflection. We use a performance metric, $P$, calculated for each participant for both systems. We denote $P_i^\text{IRDA}$ and $P_i^B$ as the performance metrics for participant $i$ on our system and the baseline system, respectively. The choice of performance metric varied between studies: Study 1 used balanced accuracy due to high class imbalance, while Study 2 used accuracy.

For each participant, we generate rewards (labels) using both systems for the 20 scenarios participants labelled that were not used for training the SL baselines. We then calculate $P$ for each system per participant. This process yields $n$ pairs of $P$ values, where $n$ is the total number of participants.
We conducted three statistical tests on the $P$ values:
\begin{enumerate}
    \item We bootstrapped 95\% confidence intervals for the mean by resampling 10,000 times with replacement.
    \item For each participant, we calculated the difference $\Delta P_i = P_i^\text{IRDA} - P_i^B$ and bootstrapped these differences in the same way.
    \item We compared the $P$ values for each system using the Wilcoxon signed-rank test. This non-parametric test was chosen over parametric alternatives as it is less prone to false positives and more robust when the data distribution is unknown or skewed \cite{bridge1999increasing}.
\end{enumerate}

This analysis primarily addresses \textbf{RQ2} by comparing the performance of systems with and without dialogic reflection, allowing us to assess if structured reflection enhances language-based reward modelling.

\subsubsection{Comparison to Supervised Learning}\label{section:comp_sl}

We compared our language-based systems to traditional supervised learning approaches. Both the individual models ($\text{MLP}^{\text{ind}}_i$ and $\text{CNN}^{\text{ind}}_i$) and the collective models ($\text{MLP}^{\text{col}}$ and $\text{CNN}^{\text{col}}$) were trained incrementally, gradually increasing the number of samples used per participant. This methodology allowed us to analyze how model performance evolved with increasing data availability.
For each increment, we calculated $P^\text{ind}_i$ and $P^\text{col}_i$ for each participant $i$, representing the performance of the individual and collective models, respectively. To ensure robust statistical analysis, we bootstrapped these values with replacement using 10,000 resamples.

This comparison helps answer \textbf{RQ3} by revealing when language-based methods outperform supervised learning approaches under various conditions. It also informs \textbf{RQ1} by showing if individualized language-based models consistently capture personal value definitions better than collective supervised models.

\subsubsection{Qualitative Analysis of Participant Decision Making}\label{section:qual_decision}
We conducted a detailed analysis of the message exchanges between participants and the system to gain insight into participants' decision-making processes. We employed an inductive coding approach, systematically reviewing the messages to identify key features and criteria that participants used in their decision-making. Our coding process involved multiple passes through the data, with iterative refinement of the codebook to ensure it captured the full range of decision-making strategies observed.

This analysis directly addresses \textbf{RQ1} by identifying different features and criteria used by participants, providing rich evidence of how value definitions differ.

\subsubsection{Analysis of Feature Similarity Between Participants}\label{section:feature_similarity}
To quantify how similar participants were in their use of decision-making features, we employed the Jaccard similarity coefficient. This measure calculates the overlap between two sets of items which, in our case, are features the two participants used to make decisions~\cite{jaccard1912distribution}. We computed the Jaccard similarity coefficient for every possible pair of participants, using the set of decision-making features each participant employed (as identified in our qualitative analysis). To estimate the overall similarity across our participant pool, we then calculated the mean of these pairwise Jaccard coefficients. To ensure robustness, we used bootstrapping with 10,000 resamples (sampling with replacement) to determine the 95\% confidence interval for this mean Jaccard similarity coefficient. 

This quantitative measure helps answer \textbf{RQ1} by providing a numerical representation of how much participants' decision-making features overlap. Low similarity scores strongly support the notion that value definitions vary significantly between participants. It also informs \textbf{RQ3} by providing a numerical indicator for when individualized vs. collective approaches might be more appropriate.

\subsubsection{Thematic Analysis of Interview Data}\label{section:thematic}
We conducted semi-structured interviews with participants to understand their experiences. The interview transcripts were analyzed using a thematic analysis approach guided by the principles outlined by Braun and Clarke \cite{braun2006using}. We followed a six-phase process: familiarization with the data, generating initial codes, searching for themes, reviewing themes, defining and naming themes, and producing the report.

This analysis addresses \textbf{RQ2} by revealing if reflection enhanced participants' ability to articulate their values, which is crucial for effective language-based reward modelling. It also informs \textbf{RQ3} as language-based reward modelling requires that participants can articulate their value definition to be effective. 

\section{Results: Study 1 - Multi-Agent Apple Farming}

% \textit{\textbf{Environment \& Task.}}
% % \subsubsection*{Design}

% % Participants first went through the dialogue phase and then labelled 50 trajectories.

% \textit{\textbf{Participants.}}
% % \subsubsection*{Participants}

% \textit{\textbf{Grid World Environment: }}

% To answer the second question, we measure how much the reward models produced by each pipeline agree with the participants they are modelling. We then compare this to how much the participants agree with one another. Specifically, we created two reward models for each participant: one using the baseline pipeline and another using our pipeline (as described in section \ref{partic_reward_modelling}). We assess the agreement between each participant's responses and their corresponding reward models using the Fleiss' kappa metric. By averaging these agreement scores across all participants for both pipelines, we obtain a measure of how well the reward models from each pipeline match the participants' views. We also bootstrap the agreement scores to obtain confidence intervals for the average agreement between the participants and the reward models from each pipeline. We then calculate the inter-annotator agreement between all participants again using Fleiss' kappa. \todo{add citation and explanation for the choice of Fleiss' kappa. Maybe add a word of warning?}

% \subsection{S1 -- Quantitative Results}

On average, participants took 15 minutes 57 seconds (SD $=$ 6 min. 43 sec., range: 6 min. 59 sec. - 30 min. 55 sec.) to complete the dialogue with the system and 13 minutes 37 seconds (SD $=$ 3 min. 2 sec., range: 6 min. 55 sec. - 18 min. 26 sec.) to complete the labelling of 50 trajectories. Of 21 participants, 7 (33.$\bar{33}\%$) entered the \textit{preference clarification loop} for one iteration.

\subsubsection{S1 - Inter-Annotator Agreement}
We observed a Fleiss' kappa value between all participants' labels on the 50 labelled trajectories of $\kappa = 0.336$, indicating ``fair'' agreement among participants \cite{landis1977measurement}. The Fleiss' kappa statistic of 0.336 we observed lends credence to the idea that human values and preferences are subjective and personal.

\subsubsection{S1 -- Evaluation of Language-Based Reward Model Performance.}
\begin{figure*}
  \centering
  \begin{minipage}[b]{0.5\textwidth}
    \centering
    \includegraphics[width=\textwidth]{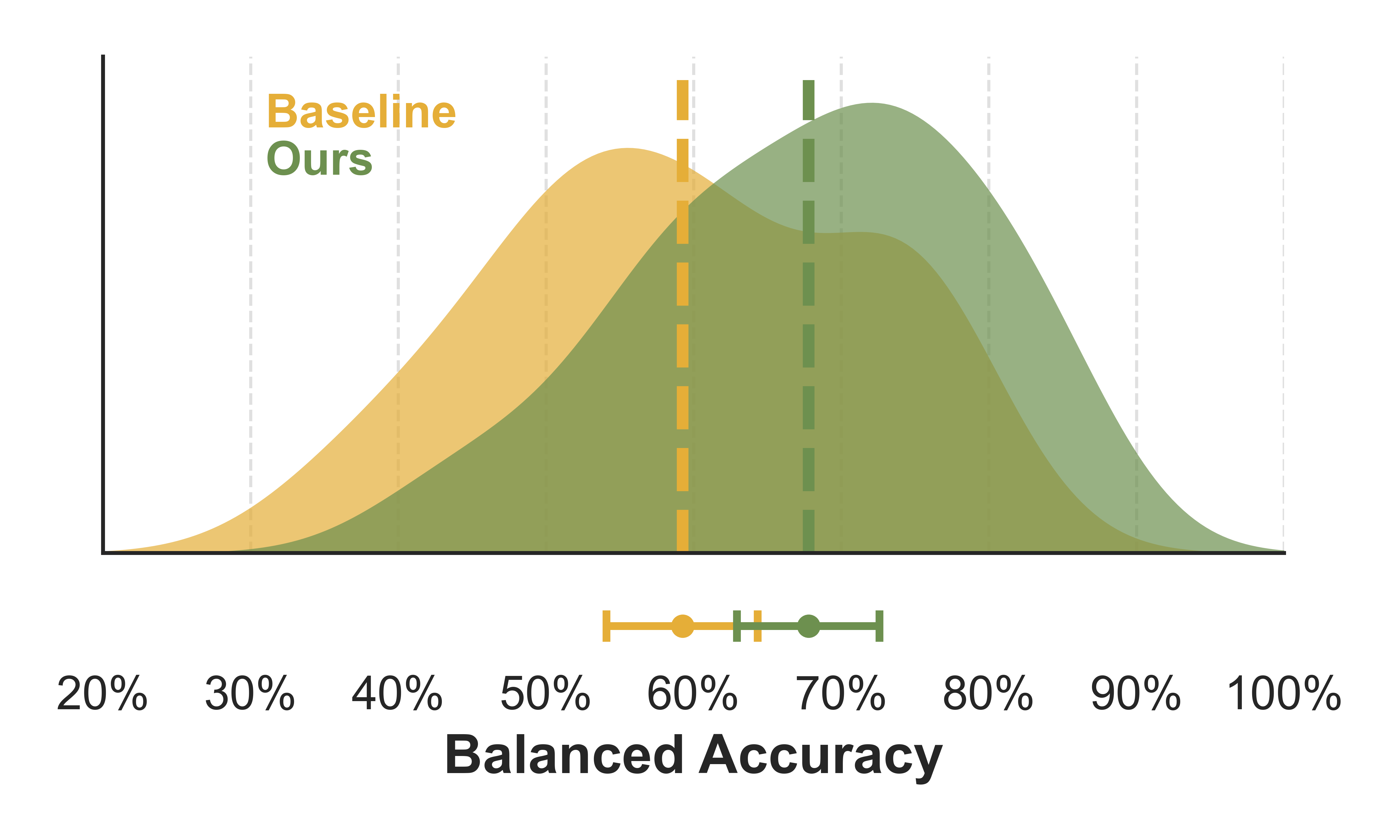}
  \end{minipage}%
  \begin{minipage}[b]{0.5\textwidth}
    \centering
    \includegraphics[width=\textwidth]{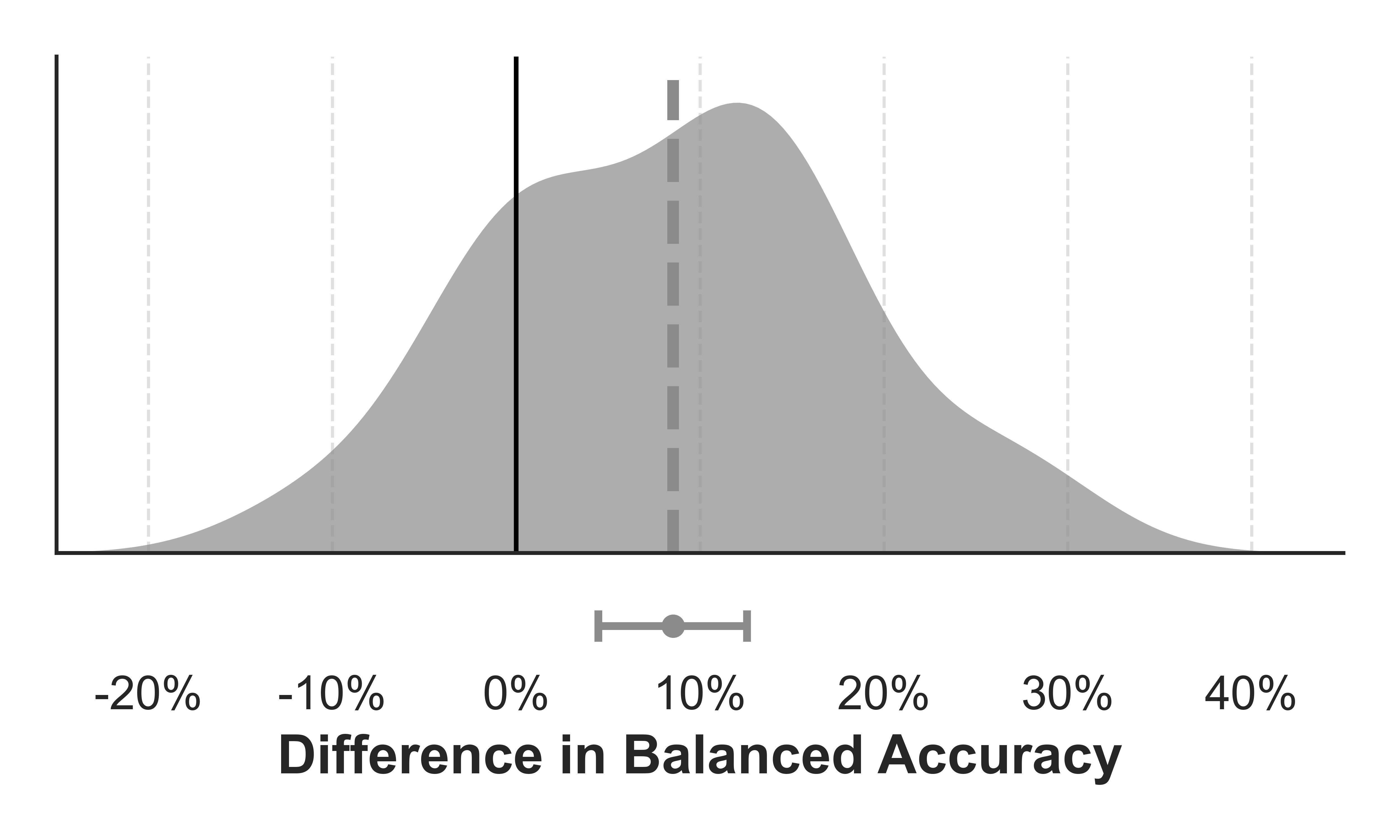}
  \end{minipage}
  \caption{(Left) Distributions of balanced accuracies for language-based reward models: our pipeline (IRDA) vs. baseline ($L^B$) in Study 1. (Right) Distribution of per-participant differences in balanced accuracy ($P^\text{IRDA}_i - P^B_i$) between IRDA and baseline models in Study 1.}
  \label{fig:accuracy_comparison_s1}
  \Description{Comparison of language-based reward model accuracies in Study 1. Left panel shows overlapping distributions of balanced accuracies for our pipeline (IRDA) and the baseline (LB), with IRDA generally showing higher balanced accuracy. Right panel displays the distribution of per-participant accuracy differences (PIRDA - PB) between IRDA and baseline models, centered slightly above zero, indicating a modest overall improvement with IRDA. Both panels include mean values (dashed lines) and confidence intervals.
}
\end{figure*}

On average, the reward models produced by our pipeline (IRDA) received significantly higher balanced accuracy scores (measured in percentages) than the baseline system ($L^B$) by 9\% (95\% CI: [5\%, 13\%], M = $68\%$ vs. M = $59\%$, p=.002). This indicates that structured reflection is beneficial.
The distributions of the balanced accuracies for each pipeline are visualized in the left frame of \autoref{fig:accuracy_comparison_s1}, and the distribution of the per participant difference in balanced accuracy is shown in the right frame.

% The average Fleiss' kappa statistic for the reward models generated with the baseline system and the participants was $\kappa = 0.25$ (95\% CI:[0.14, 0.35]), indicating. The average Fleiss' kappa for the reward models generated with our pipeline and the participants was $\kappa = 0.36$ (95\% CI:[0.28, 0.43]).

% This result indicates that the reward models produced by our pipeline, using \systemname{}, on average, agree more with each participant than the participants agree with each other about what agent behaviour is respectful. \carter{However, this result should be taken with caution?}

% \begin{figure}
%   \centering
%   \includegraphics[width=0.5\linewidth]{figures/balanced_accuracy_s1.png}
%   \caption{Distributions of the balanced accuracies of the language-based reward models generated with our pipeline (IRDA) compared to those of the language-based reward model generated with the baseline pipeline in Study 1.}
%   \Description{}
%   \label{accuracy_dist_s1}
% \end{figure}

% \begin{figure}
%   \centering
%   \includegraphics[width=0.5\linewidth]{figures/balanced_accuracy_diff_s1.png}
%   \caption{Distribution of the per participant difference in balanced accuracy between the language-based reward model generated with our pipeline (IRDA) compared to the language-based reward model generated with the baseline pipeline in Study 1.}
%   \Description{}
%   \label{accuracy_diff_s1}
% \end{figure}

\subsubsection{S1 -- Comparison to Supervised Learning}

\begin{figure}
  \centering
  \includegraphics[width=0.5\textwidth]{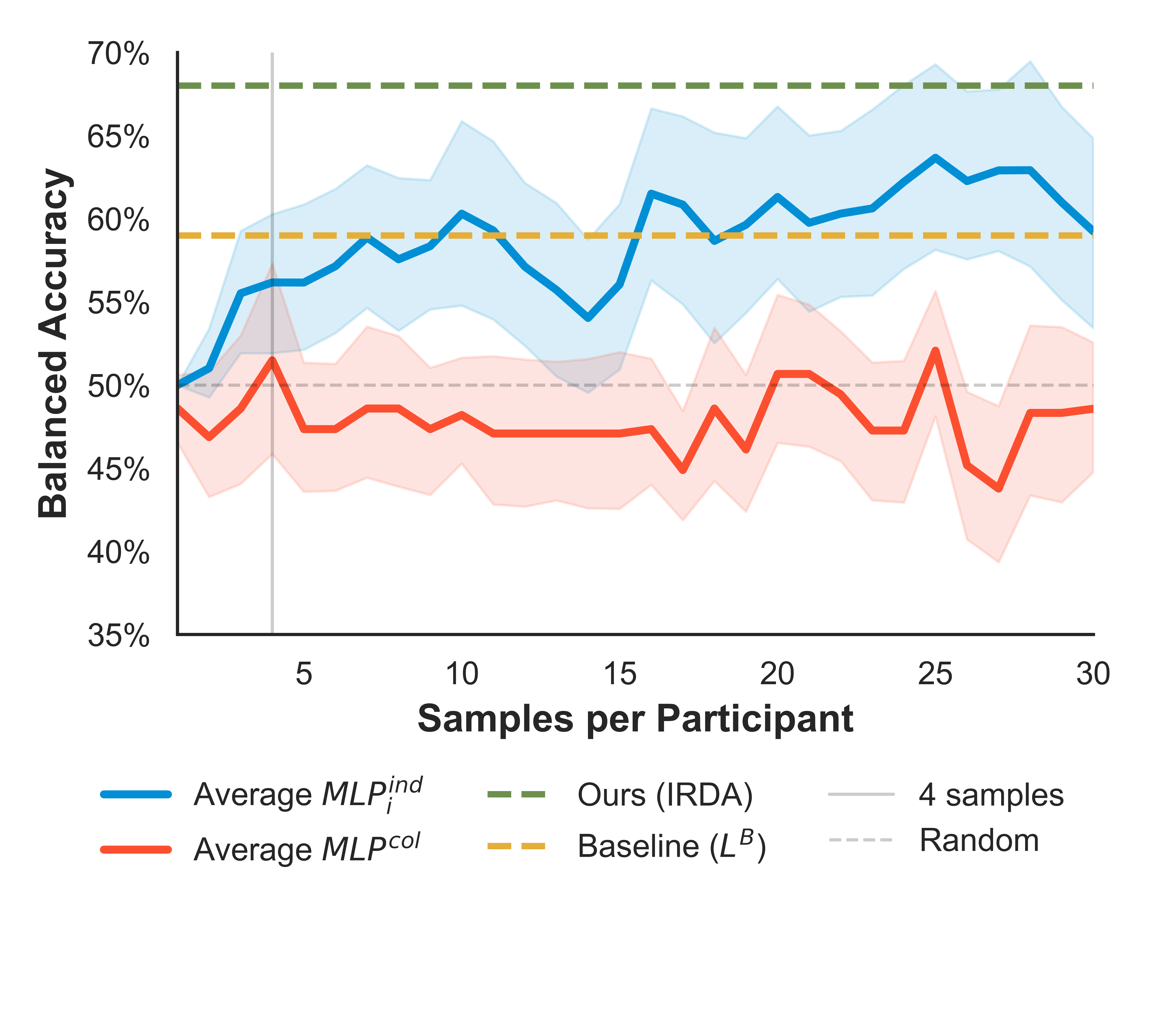}
  \caption{Balanced accuracy of different models as a function of samples per participant in Study 1. The blue line represents the average individual MLP model (MLP\textsuperscript{ind}), while the red line shows the collective MLP model (MLP\textsuperscript{col}). Our proposed system (IRDA, green dashed line) and the baseline ($L^B$, yellow dashed line) were trained on 4 samples per participant, as indicated by the vertical gray line. The collective model was trained on 21 times the number of samples shown on the x-axis due to the 21 participants in Study 1. Shaded areas represent 95\% confidence intervals. The gray dashed line indicates random performance.}
  \Description{ Line graph comparing balanced accuracy of different reward modeling approaches as sample size increases. The individual MLP model (blue line) shows a general upward trend, outperforming the collective MLP model (red line) consistently. Our IRDA system (green dashed line) maintains the highest accuracy throughout, while the baseline (yellow dashed line) remains above the collective model but below IRDA. Notably, the individual MLP model's performance improves with more samples, approaching but not surpassing IRDA's accuracy even at 30 samples per participant. The collective model's performance remains relatively stable, showing less improvement with increased samples. Confidence intervals (shaded areas) narrow as sample size increases, indicating greater certainty in model performance.}
  \label{SL_s1}
\end{figure}

With all 30 training samples, the average balanced accuracy of the individual models ($\text{MLP}^\text{ind}_i$) was 59\% (95\% CI: [53\%, 65\%]) while the collective model ($\text{MLP}^{\text{col}}$) achieved 48\% (95\% CI: [46\%, 50\%]). This indicates that participant value definitions varied significantly.
\autoref{SL_s1} illustrates the relationship between model performance and the number of samples provided per participant.

% \subsection{S1 -- Qualitative Results}
% \todo{put something here}
\subsubsection{S1 -- Qualitative Analysis of Participant Decision Making}\label{section:s1_features}

While our system is designed to align AI agents with a range of user-defined values, our evaluation concentrated on the singular value of respect. Focusing on a single value limited the variability in the study and allowed us to assess the personal and subjective nature of even a single value.

We analyzed the users' conversations with the system in Study 1 to understand what each participant thought respectful behaviour in this environment was. We found 12 behavioural features that participants used to explain whether they thought the agent was respectful. Interestingly, P1 only used one behavioural feature, namely whether the agent stayed in the quadrant it owned, to determine whether the agent acted respectfully. In contrast, P5, P7, and P10 explained whether the agent acted respectfully using seven features. Overall, only one pair of participants used the same features to make their decisions. Further, participants combined the features in diverse ways, such as forming hierarchies of importance or combining them conditionally. The 12 features we identified are:
\begin{enumerate}
    \item \textbf{Stays in Own Quadrant:} Whether the main agent stays in the orchard (quadrant) they own.
    \item \textbf{Interferes With Others:} This feature evaluates a relatively wide range of behaviours, including if the agent appeared to attempt to block another agent, following another agent, or getting too close to another agent.
    \item \textbf{Task Completion:} Whether the agent works toward or completes the task the participant thought the agent should be doing. For example, an agent picking up all apples or garbage in their quadrant. 
    \item \textbf{Picks Up Own Garbage:} Whether the agent picks up garbage in their own orchard (quadrant)
    \item \textbf{Picks Up Others' Garbage:} Whether the agent picks up garbage in other agents orchards
    \item \textbf{Tit for Tat Behaviour: }Some participants explained that a given behaviour was respectful if another agent had done it to them first, such as entering their quadrant.
    \item \textbf{Taking Others' Apples}: Whether the agent picks up apples in orchards it does not own.
    \item \textbf{Eats Own Apples: }Whether the agent eats any apples in their own quadrant.
    \item \textbf{Picks Up Garbage Before Apple:} Whether the agent picks up garbage before eating apples in their own or other quadrants.
    \item \textbf{Efficiency: }Whether the agent moves around without collecting apples or garbage or makes repetitive, futile movements
    \item \textbf{Time in Others' Quadrants:} The duration of time spent in other agents' orchards.
    \item \textbf{In Quadrant While Owner was Gone: }If the agent entered another agent's quadrant while they were gone.
\end{enumerate}

\begin{figure*}
  \centering
  \includegraphics[width=\linewidth]{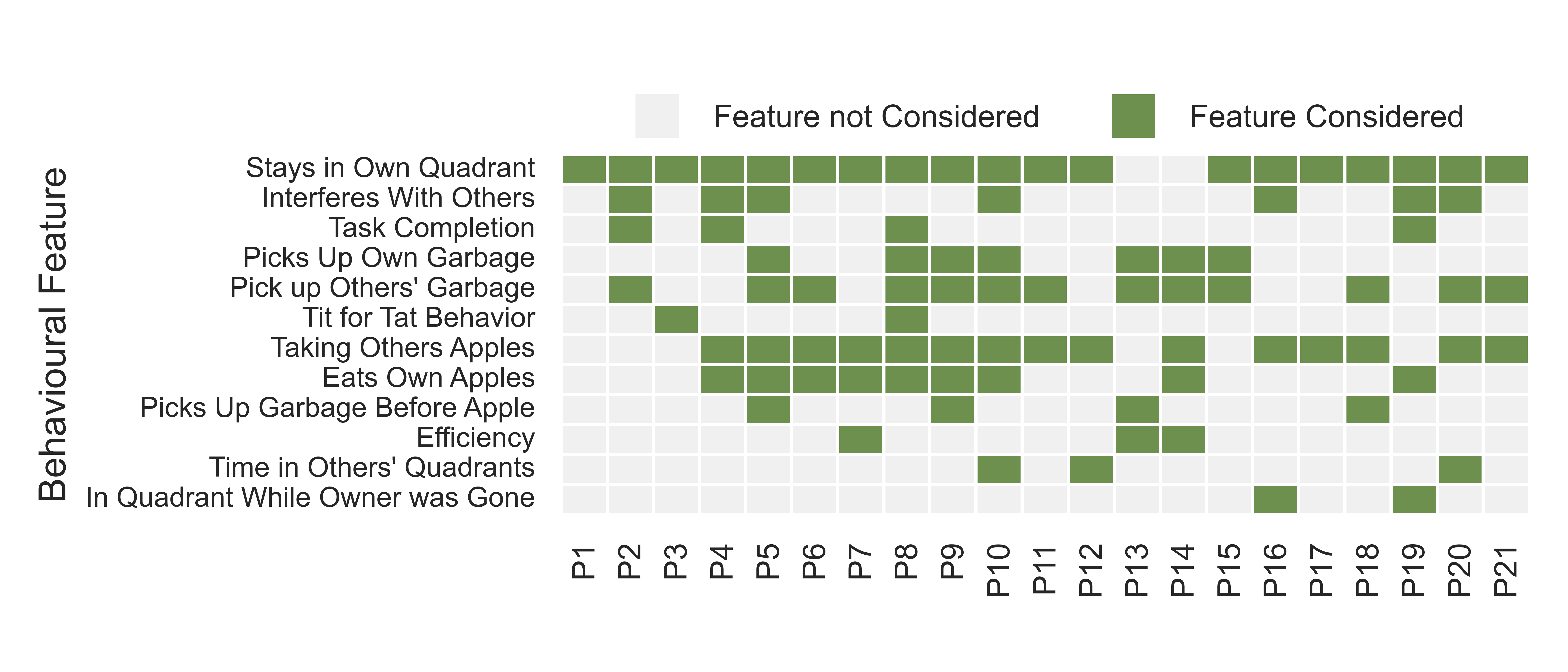}
  \caption{Behavioural features participants used to decide whether the agent was acting respectfully in Study 1.}
  \Description{Heatmap illustrating behavioural features considered by participants (P1-P21) when assessing agent respectfulness in Study 1. The chart shows 11 features, with green cells indicating a feature was considered and gray cells indicating it was not. "Stays in Own Quadrant" is the most consistently considered feature across participants, while "In Quadrant While Owner was Gone" is least considered. There's significant variability in feature consideration among participants, with some (e.g., P1, P5) considering many features and others (e.g., P13, P15) focusing on fewer. This visualization reveals diverse individual approaches to evaluating agent behaviour.
}
  \label{respect_features}
\end{figure*}

Among the features, the most commonly used was \textbf{Stays in Own Quadrant}, but participants often found other features more important. Some features are temporally static, while others span multiple steps. For example, whether an agent is in its quadrant can be determined at a single time, but whether an agent picked up garbage before eating an apple must be determined over multiple steps. The features each participant used are shown in \autoref{respect_features}.

\subsubsection{S1 -- Analysis of Feature Similarity Between Participants. }

We observed an average Jaccard similarity coefficient between all pairs of participants' feature usage of $J = 0.357$, with a 95\% confidence interval of $(0.333, 0.3813)$.

\subsubsection{S1 -- Thematic Analysis of Interview Data} Through our thematic analysis, we found two main themes: (1) participants' definitions of respect evolved throughout the activity, and (2) The system's hypothesis and alternative perspectives had a specific impact on this evolution.
% Through our thematic analysis, we identified a number of themes, \todo{}
 \space
 
\textit{\textbf{Evolving Definitions of Respect.}}
% \subsubsection*{}
The interview data illustrates that participants' understandings of respect evolved significantly through interaction with the system, particularly influenced by exposure to examples and the system's alternative features. This finding aligns with previous work in consumer research, which found that engaging users in reflection and allowing them to make realistic decisions increases the accuracy of their reported preferences~\cite{hauser2014self}. Initially, many participants held simplistic definitions of respect, often related to spatial boundaries or specific tasks like collecting one's own apples (P4, P6, P7, P10, P19). However, these definitions became more nuanced as participants engaged with the system. The evolution was often spurred by witnessing examples that challenged their initial views and contemplating alternative definitions presented by the system (P3, P8, P10, P13, P18, P19, P20). This highlights a process where the system's feedback and the act of labelling examples prompted participants to refine and sometimes expand their conceptualization of respect, moving beyond their initial assumptions.

\textit{\textbf{Specific Impact of System Hypothesis.}}
% \subsubsection*{Impact of System Hypothesis}
Another recurrent theme was the specific influence of the system's hypotheses and feedback on shaping participants' conceptions of respect. Participants reported that the system's presentation of alternative features and hypotheses prompted reevaluation, clarification and, in some cases, a significant revision of their definitions of respect (e.g., P3, P8, P13, P18, P20). For instance, the system's suggestions helped participants to narrow down their considerations (P13, P18) or to think about respect in ways they had not initially contemplated (P3, P20). While the system's alternative features did not always change participants' minds (e.g., P19), they played a role in the iterative process of refining participants' understandings, illustrating the value of engaging the user in reflective dialogue.

\section{Results: Study 2 - The Moral Machine}

On average, participants took 18 minutes 28 seconds (SD $=$ 7 min. 15 sec., range: 12 min. 0 sec. - 34 min. 34 sec.) to complete the dialogue with the system and 11 minutes 51 seconds (SD $=$ 4 min. 15 sec., range: 4 min. 46 sec. - 17 min. 04 sec.) to complete the labelling of 50 trajectories. Of 9 participants, 1 (11.$\bar{11}\%$) entered the \textit{preference clarification loop} for one iteration.

\subsubsection{S2 - Inter-Annotator Agreement}
We observed a Fleiss' kappa value between all participants' labels on the 50 labelled trajectories of $\kappa = 0.460$, indicating ``moderate'' (higher than ``fair'') agreement among participants \cite{landis1977measurement}.

\subsubsection{S2 -- Evaluation of Language-Based Reward Model Performance.}
\begin{figure*}
  \centering
  \begin{minipage}[b]{0.5\textwidth}
    \centering
    \includegraphics[width=\textwidth]{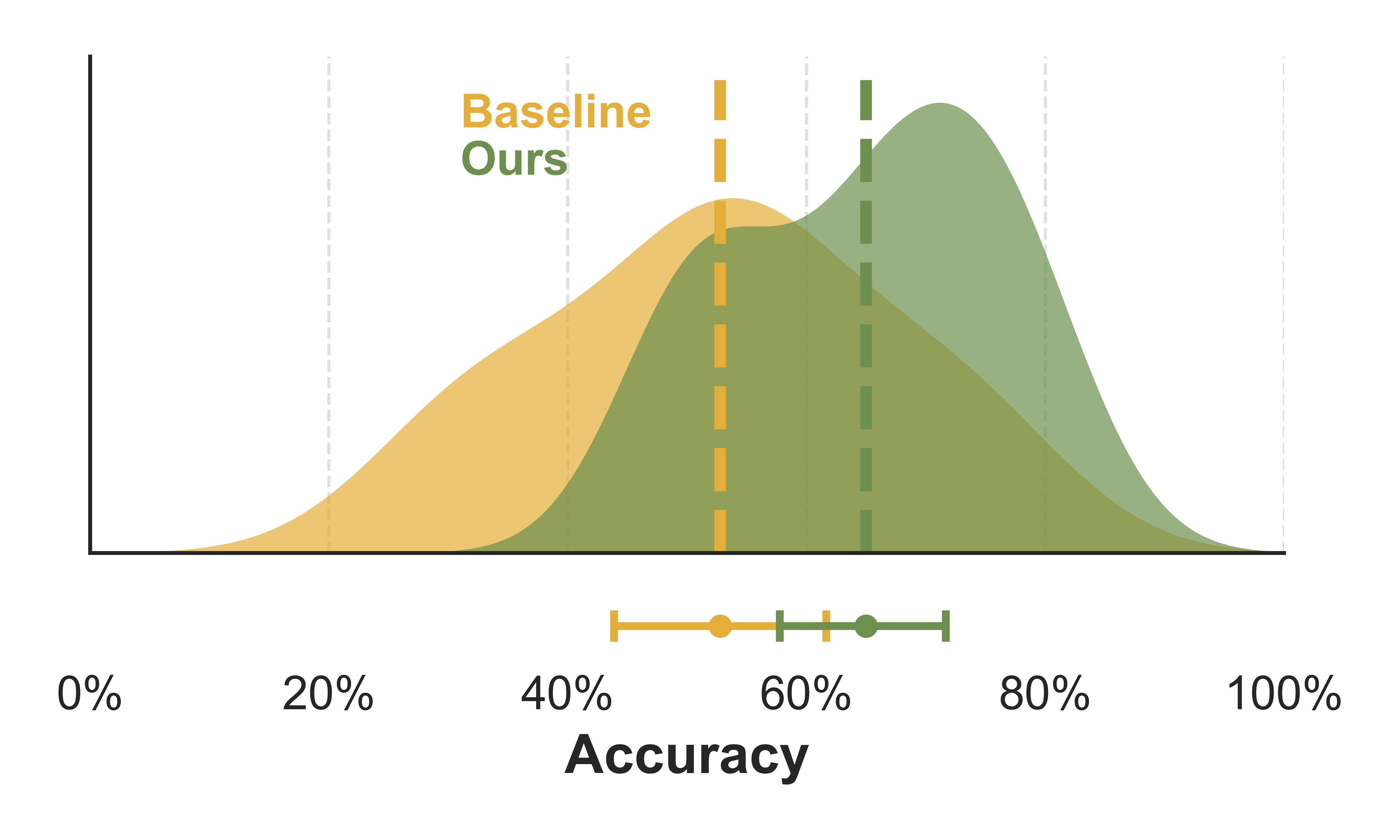}
  \end{minipage}%
  \begin{minipage}[b]{0.5\textwidth}
    \centering
    \includegraphics[width=\textwidth]{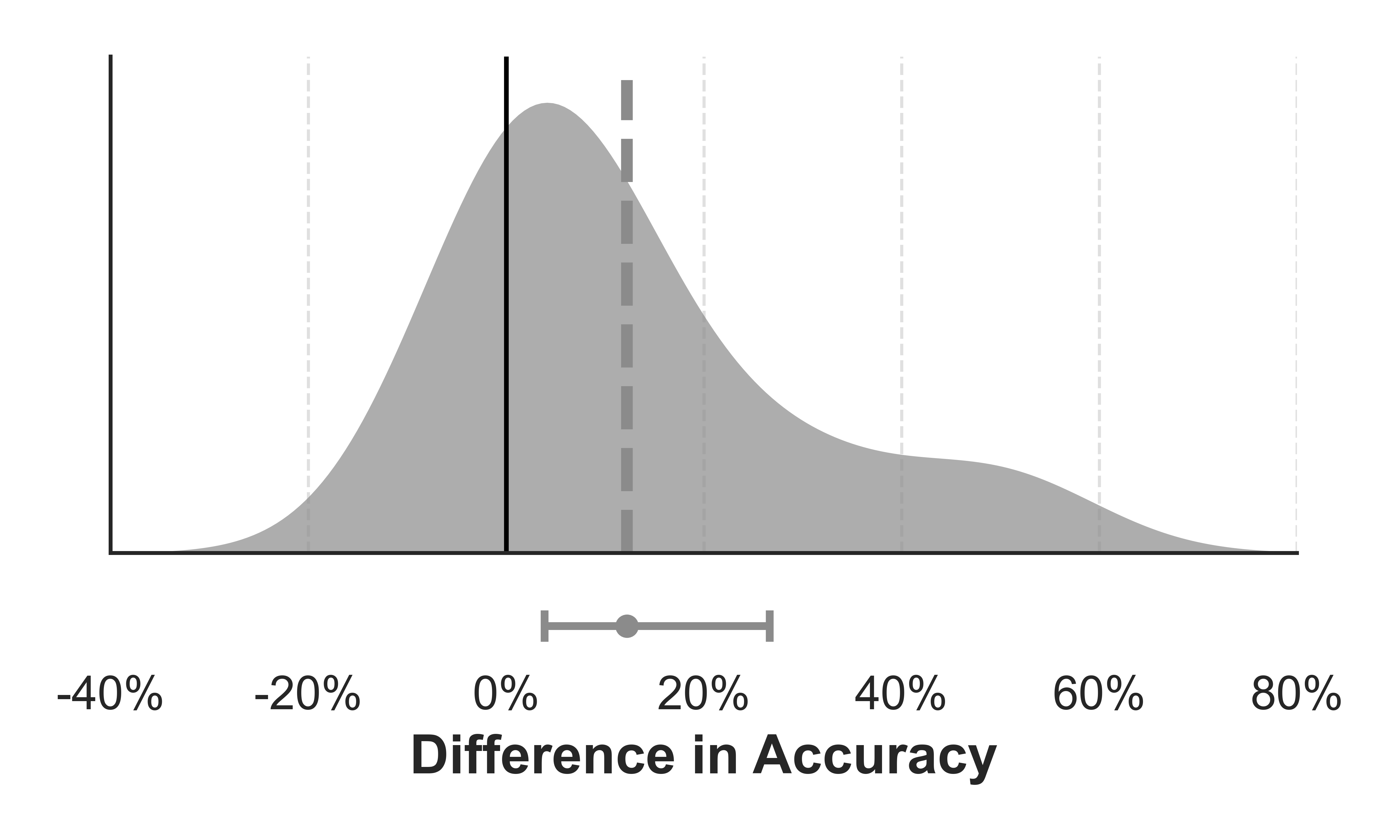}
  \end{minipage}
  \caption{(Left) Distributions of accuracies for language-based reward models: our pipeline (IRDA) vs. baseline in Study 2. (Right) Distribution of per-participant differences  in accuracy ($P^\text{IRDA}_i - P^B_i$) between IRDA and baseline models in Study 2.}
  \Description{Comparison of language-based reward model accuracies in Study 2. Left panel shows overlapping distributions of accuracies for our IRDA pipeline (green) and the baseline (yellow), with IRDA demonstrating higher mean accuracy and a broader distribution towards higher values. Right panel displays the distribution of per-participant accuracy differences between IRDA and baseline models, centered slightly above zero with a positive skew, indicating IRDA's overall superior performance. Both panels include mean values (dashed lines) and confidence intervals, illustrating the statistical significance of IRDA's improvement over the baseline.}
  \label{fig:accuracy_comparison_S2}
\end{figure*}

On average, the reward models produced by our pipeline (IRDA) received significantly higher accuracy scores (measured in percentages) than the baseline system ($L^B$) by 12\% (95\% CI: [4\%, 27\%], M = $65\%$ vs. M = $53\%$, p=.05). This adds more evidence in favour of the effectiveness of structured reflection. The distributions of the balanced accuracies for each pipeline are visualized in the left frame of \autoref{fig:accuracy_comparison_S2}. The distribution of the per-participant difference in balanced accuracy is shown in the right frame.

% The average Fleiss' kappa statistic for the reward models generated with the baseline system and the participants was $\kappa = 0.25$ (95\% CI:[0.14, 0.35]), indicating. The average Fleiss' kappa for the reward models generated with our pipeline and the participants was $\kappa = 0.36$ (95\% CI:[0.28, 0.43]).

% This result indicates that the reward models produced by our pipeline, using \systemname{}, on average, agree more with each participant than the participants agree with each other about what agent behaviour is respectful. \carter{However, this result should be taken with caution?}

% \begin{figure}
%   \centering
%   \includegraphics[width=0.5\linewidth]{figures/accuracy_comparison_S2.png}
%   \caption{Distributions of the accuracies of the language-based reward models generated with our pipeline (IRDA) compared to those of the language-based reward model generated with the baseline pipeline in Study 2.}
%   \Description{}
%   \label{accuracy_dist_S2}
% \end{figure}

% \begin{figure}
%   \centering
%   \includegraphics[width=0.5\linewidth]{figures/accuracy_diff_S2.png}
%   \caption{Distribution of the per participant difference in accuracy between the language-based reward model generated with our pipeline (IRDA) compared to the language-based reward model generated with the baseline pipeline in Study 2.}
%   \Description{}
%   \label{accuracy_diff_S2}
% \end{figure}

\subsubsection{S2 -- Comparison to Supervised Learning}

% In this study, we trained two types of reward models with supervised learning: a simple MLP and a Convolutional Neural Network. \todo{redundant} One version is based on a simple MLP network with one hidden layer of 32 neurons and uses a highly processed (feature-engineered) input. The other is a convolutional neural network (CNN) \todo{Put details in appendix} consisting of two convolutional layers, each followed by a Rectified Linear Unit (ReLU) activation function and max pooling operation. These initial layers are responsible for extracting and refining visual features from the input images. Following the convolutional layers, the network uses a flattening operation to convert the 2D feature maps into a 1D vector. This flattened representation is then fed into a fully connected layer with ReLU activation, which further processes the extracted features. The network culminates in a final output layer with two neurons, corresponding to the two classes in the binary classification task. 

% The Moral Machine setting lends itself to an easy \todo{featurization} (i.e., the scenarios can be succinctly described numerically), and we trained the MLP on this. For the CNN, we used images of the moral machine scenarios. Since the CNN must learn the relevant features in the convolutional layers, as opposed to being provided them, the learning problem is more difficult. 

\begin{figure*}
  \centering
  \begin{minipage}[b]{0.5\textwidth}
    \centering
    \includegraphics[width=\textwidth]{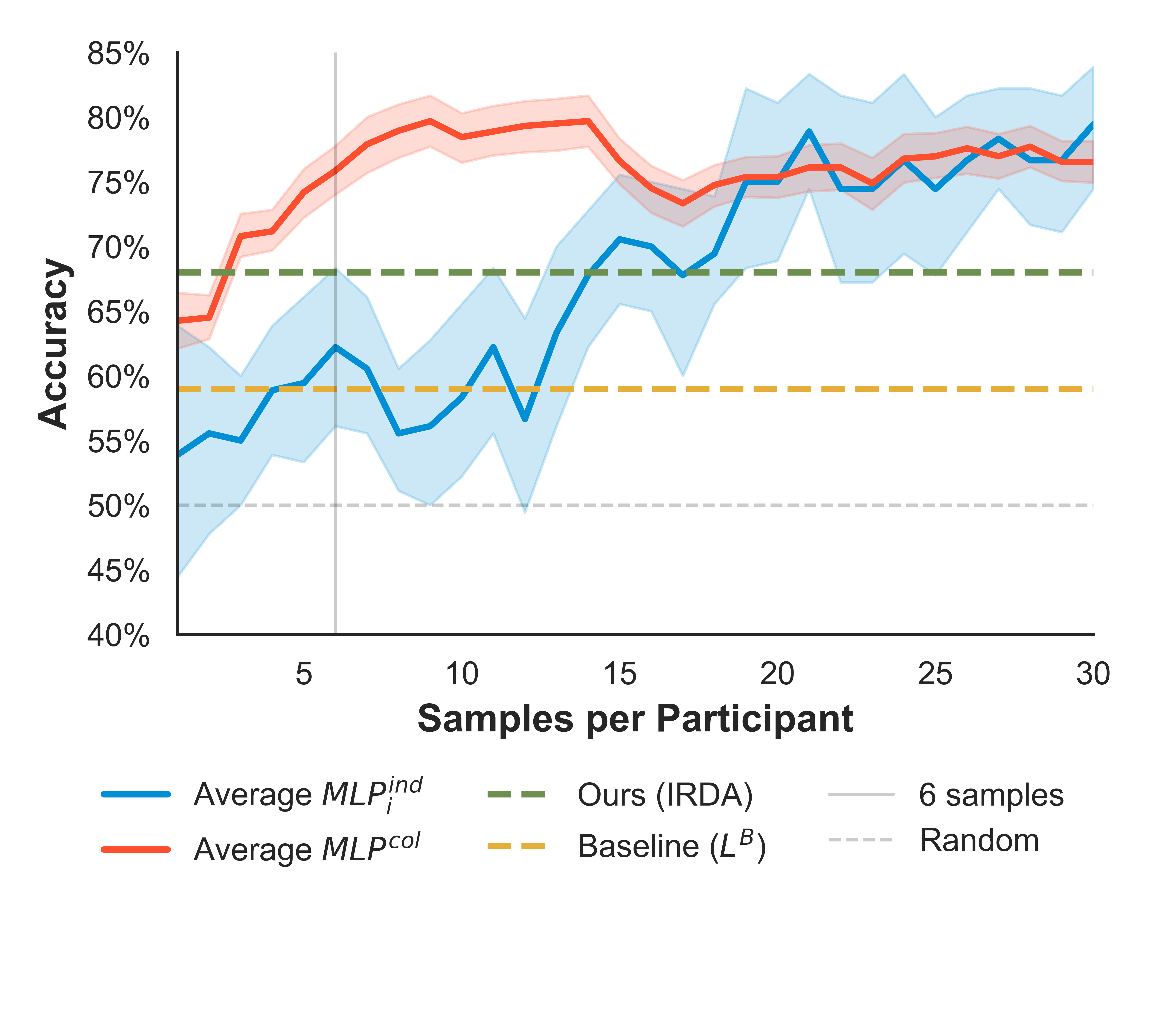}
  \end{minipage}%
  \begin{minipage}[b]{0.5\textwidth}
    \centering
    \includegraphics[width=\textwidth]{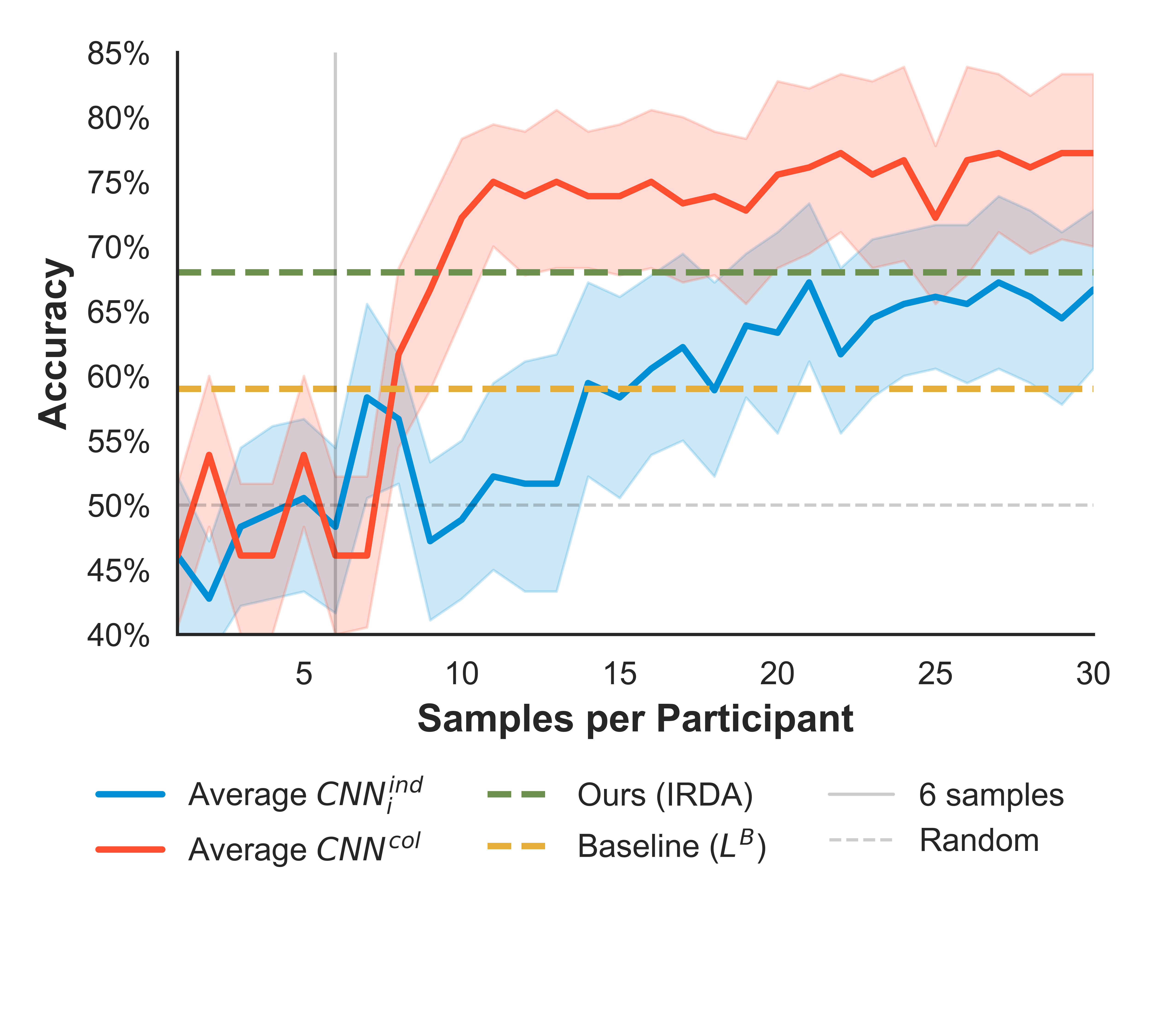}
  \end{minipage}
  \caption{Comparison of model accuracies as a function of samples per participant in Study 2.
(Left) Performance of MLP-based models: average individual MLP (MLP\textsuperscript{ind}, blue), collective MLP (MLP\textsuperscript{col}, red), our IRDA approach (green dashed), and baseline $L^B$ (yellow dashed).
(Right) Performance of CNN-based models: average individual CNN (CNN\textsuperscript{ind}, blue), collective CNN (CNN\textsuperscript{col}, red), IRDA, and baseline.
Both panels show confidence intervals (shaded areas), the 6-sample training point for IRDA and baseline (vertical gray line), and the random performance level (gray dashed). IRDA consistently outperforms other approaches across model architectures.}
  \Description{Comparison of MLP-based (left) and CNN-based (right) model accuracies as sample size increases. The collective MLP model quickly outperforms IRDA (at around 3 x 9 = 27 samples), while the collective CNN surpasses it at about 9x9=81 samples). Both collective models maintain their lead thereafter. Individual models show slower improvement, with the MLP version eventually surpassing IRDA at approximately 17 samples, whereas the individual CNN model never exceeds IRDA's performance within the observed range. IRDA consistently outperforms the baseline (LB) across all sample sizes for both architectures. }
  \label{fig:sl_s2_comparison}
\end{figure*}

With all 30 training samples, the average accuracy of the individual MLP models ($\text{MLP}^\text{ind}_i$ was 79\% (95\% CI: [74\%, 84\%]) while the collective model ($\text{MLP}^{\text{col}}$) achieved 77\% (95\% CI: [75\%, 78\%]). The left frame \autoref{fig:sl_s2_comparison} illustrates the relationship between model performance and the number of samples provided per participant for the MLP models. For the CNN models, with all 30 training samples, the average accuracy of the individual models ($\text{CNN}^\text{ind}_i$) was 67\% (95\% CI: [61\%, 73\%]) while the collective model ($\text{CNN}^{\text{col}}$ achieved 77\% (95\% CI: [70\%, 83\%]). The right frame of \autoref{fig:sl_s2_comparison} illustrates the relationship between model performance and the number of samples provided per participant for the CNN models. These results point to two things: first, when participant agreement is high, collective methods may outperform individualized methods. Second, when agreement is high and the learning problem becomes more difficult (e.g., the CNN models with image input), pooling of samples is beneficial.

% \begin{figure}
%   \centering
% \includegraphics[width=0.5\textwidth]{figures/sl_S2_cv_CNN.png}
%   \caption{\todo{}}
%   \Description{}
%   \label{SL_S2_cv_CNN}
% \end{figure}

% \subsection{S2 -- Qualitative Results}

\subsubsection{S2 -- Qualitative Analysis of Participant Decision Making}\label{section:s2_features}

\begin{figure}
  \centering
  \includegraphics[width=0.98\linewidth]{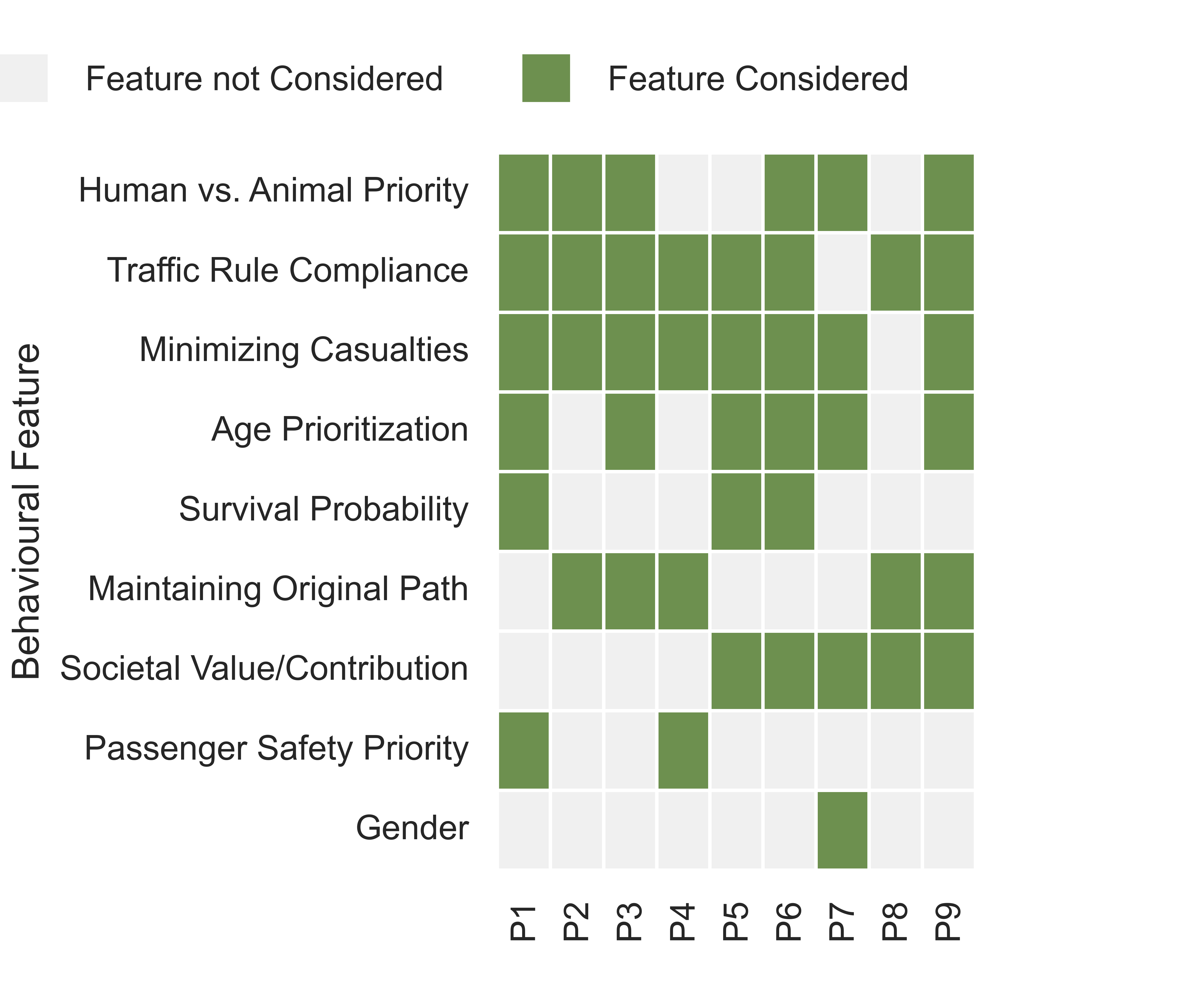}
  \caption{Behavioural features participants used to decide what the autonomous vehicle should do in Study 2.}
  \Description{Heatmap illustrating behavioral features considered by participants (P1-P9) when evaluating autonomous vehicle decision-making. The chart displays nine ethical considerations, with green cells indicating a feature was considered and gray cells indicating it was not. "Traffic Rule Compliance" and "Minimizing Casualties" are the most consistently considered features across participants, while "Gender" is least considered. There's notable variability in feature consideration among participants, with some considering a wide range of factors and others focusing on fewer. }
  \label{features_S2}
\end{figure}

By analyzing the participants' conversations with our system, we identified nine features they used in their decision-making in Study 2. The features we identified are:
\begin{enumerate}
    \item \textbf{Human vs. Animal Priority:} Whether human lives are prioritized over animal lives.
    \item \textbf{Traffic Rule Compliance:} The consideration of the legality of pedestrians' actions.
    \item \textbf{Minimizing Casualties:} The aim to minimize the total number of fatalities.
    \item \textbf{Age Prioritization:} The preference for saving younger people over older people.
    \item \textbf{Survival Probability:} The consideration of the likelihood of survival for different individuals based on factors like physical fitness.
    \item \textbf{Maintaining Original Path:} The preference for the vehicle to stay on its original course rather than swerving.
    \item \textbf{Societal Value:} The consideration of perceived societal value or potential future contributions of individuals.
    \item \textbf{Passenger Safety Priority:} The prioritization of the safety of the vehicle's passengers over pedestrians or other road users.
    \item \textbf{Gender:} The consideration of the gender of potential victims in the decision-making process.
\end{enumerate}

Like Study 1, participants combined and used these features in various ways.

\subsubsection{S2 -- Analysis of Feature Similarity Between Participants. }

We observed an average Jaccard similarity coefficient between all pairs of participants' feature usage of $J = 0.464$, with a 95\% confidence interval of $(0.403, 0.526)$.

\subsubsection{S2 -- Thematic Analysis of Interview Data}Through our thematic analysis, we found two main themes: (1)
participants’ definitions of respect evolved throughout the activity, and (2) participants' decisions were largely based on explicit reasoning but sometimes relied on intuition.

\textit{\textbf{Decision-making Evolution. }}
Our results reveal a divergence in how participants' decision-making processes evolved throughout the study. Some participants reported that their approach changed as they encountered a wider range of scenarios. For instance, P3 noted that "as more cases came up, I realized the need to consider new factors when the initial factors were equal between groups." This reiterates our finding from Study 1 that exposure to diverse situations can prompt users to refine and expand their decision-making criteria. Conversely, other participants maintained consistent rules throughout the study. P2 stated that their "rules remained consistent throughout," indicating that some users may approach such tasks with pre-established ethical frameworks that remain stable across various scenarios. 
% This divergence highlights the importance of considering individual differences in ethical reasoning when designing AI systems that involve human-in-the-loop decision making.

% \textit{\textbf{Impact of AI Feedback. }}
% Compared to Study 1, participants were less influenced by the reflective dialogue offered by the system. While P5 reported changing their answers based on the reflective dialogue offered by the system and entered the preference clarification loop, no other participants entered the preference clarification loop.

\textit{\textbf{Intuition vs. Explicit Reasoning. }}
Our findings reveal an interplay between explicit reasoning and intuition. Most participants felt capable of articulating their decision-making process. Still, the emergence of intuition-based decisions in particularly challenging scenarios, as reported by P6 and P7, underscores the complexity of ethical reasoning. P7 mentioned relying on "first instinct" or "vibes" for 3 or 4 especially difficult scenarios. 
% This finding has important implications for the design of ethical AI systems. While promoting explicit reasoning is generally desirable, interfaces should also acknowledge and potentially accommodate intuitive decision-making in complex cases. Future research could explore how to balance these two modes of thinking in AI-assisted ethical decision-making tools.
% Value of Scenario Exposure
% The study underscores the importance of exposing users to a diverse range of scenarios when engaging with ethical decision-making systems. Several participants, including P3 and P4, emphasized that encountering different scenarios helped them develop or refine their decision-making process. This suggests that scenario-based learning could be an effective approach for training users to interact with AI systems that handle ethical dilemmas.

% \textit{\textbf{Comprehensiveness of Explanations. }}
% Most participants felt that their explanations during the dialogue phase captured a significant portion (90\% or more) of their decision-making process. This finding, reported by P6 and P7, suggests that the dialogue-based approach was largely successful in eliciting comprehensive explanations from users.

\section{Discussion}
Our evaluation of \systemname{} yields insights that challenge fundamental assumptions about human values and AI alignment. These findings have implications for the design of AI systems that aim to respect and represent diverse values and preferences.

\subsection{The Diversity of Human Values (RQ1)}

In both Study 1 and 2, we observed heterogeneity in participants' value definitions. However, participants' value definitions varied to a greater degree in Study 1. 

The diversity in participants' interpretations of respectful behaviour, observed in Study 1, suggests that values like respect are not universal constants but rather individualized constructs. This heterogeneity manifested in several ways: The identification of 12 distinct behavioural features used by participants to evaluate respectful agent behaviour reveals the multifaceted nature of respect. Further, the unique combinations of these features used by participants highlight the individualized nature of value interpretation. This finding extends previous work on value pluralism in moral philosophy \cite{berlin1969four}, suggesting that even seemingly universal values are subject to significant individual variation.

The ``fair'' inter-annotator agreement ($\kappa = 0.23$) and relatively low 
feature usage similarity ($J = 0.357$) in Study 1 further underscores this diversity. This level of disagreement, observed when participants evaluated identical agent behaviours, implies that discussions about values may often involve fundamentally different conceptualizations masked by shared terminology.

Perhaps the most compelling evidence for this diversity in Study 1 comes from comparing the performance of individual models ($\text{MLP}^\text{ind}_i$) with the collective reward model ($\text{MLP}^\text{col}$). Individual models achieved an average balanced accuracy of 59\%, demonstrating that above-random performance is possible in this task with only 30 samples. However, despite access to 630 samples, the collective reward model could not surpass random performance. This contrast suggests that the diversity in value interpretations is not simply noise that can be averaged out with more data. Instead, it indicates genuine and significant differences in how individuals interpret values.

Study 2, in comparison, revealed more homogeneous opinions, with higher inter-annotator agreement ($\kappa = 0.460$) and greater feature usage similarity ($J = 0.464$). In accordance with this, the collective models ($\text{MLP}^\text{col}$ and $\text{CNN}^\text{col}$) were generally more effective in capturing participants' value definitions than individualized models.

The contrast between Studies 1 and 2, where Study 2 revealed more homogeneous opinions, highlights the context-dependent nature of value diversity. This finding suggests that the degree of consensus about values may vary significantly across different scenarios or domains. These observations challenge the assumption that a universal set of values can be embedded in AI agents in all contexts. Given the diversity in value definitions, personalization or, in multi-stakeholder settings, making informed compromises that respect all stakeholders are both viable alignment strategies. Individualized reward models facilitate both personalization and informed and interpretable compromises.

\subsection{The Power of Structured Reflection (RQ2)}

The increased performance of our reflection-based system compared to the non-reflective language-based baseline in Studies 1 and 2 demonstrates the effectiveness of structured reflection in preference elicitation and communication. This finding suggests that guiding users to consider alternative perspectives and reflect on their preferences enables a more accurate capture of individual value definitions.

This observation aligns with dual-process theories in cognitive psychology \cite{evans2019reflections}, indicating that our reflection process engages System 2 thinking - the deliberate and analytical mode of thought. By prompting users to articulate and justify their value judgments, we facilitate a process of value discovery and refinement. 

Qualitative feedback from Study 1 participants provides insight into this process. Users reported that engaging with the system's hypotheses and alternative features allowed them to refine or clarify their understanding of respectful behaviour. This suggests that the reflection process serves as a form of cognitive scaffolding, supporting users in exploring and articulating their values. Drawing upon research on "designing for reflection" \cite{fleck2010reflecting}, our dialogue system creates opportunities for users to externalize and examine their internal value frameworks.

Intriguingly, Study 2 revealed that our system outperformed the baseline even when participants had more fixed opinions and engaged less with the preference clarification loop. This finding highlights a crucial aspect of our approach: structured reflection enhances preference communication, improving the system's ability to capture nuanced preferences even when users' fundamental views remain unchanged. This aligns with recent work that suggests reflective System 2 thinking can help individuals justify, rationalize, and explain their intuitions \cite{cecchini2021dual}. 

These results suggest that AI systems may need to do more than observe behaviour or collect binary feedback to understand and align with human values. Instead, they may need to engage humans in a process of structured reflection, creating a dialogue that helps both the human and the AI system develop a clearer understanding of the human's values and preferences. Our findings contribute to the ongoing discourse on value alignment in AI, suggesting that effective preference elicitation requires not just sophisticated AI models, but also carefully designed interaction paradigms that support human cognitive processes.

\subsection{The Contextual Efficacy of Individualized Language-Based Reward Modeling (RQ3)}

Our investigation into the efficacy of individualized language-based reward modelling (RQ3) reveals a nuanced landscape of strengths and limitations. The approach demonstrates remarkable sample efficiency, outperforming individual models trained on 30 samples and collective models trained with 630 samples with just four samples in Study 1. This efficiency stems from our method's structured reflection process and the few-shot learning capabilities of LLMs.
However, this efficiency is context-dependent, with effectiveness varying based on preference heterogeneity within the population. Study 1, characterized by higher preference diversity (Fleiss' kappa = 0.336, Jaccard similarity = 0.357), showcased the strengths of our individualized approach. In contrast, Study 2 exhibited more homogeneous preferences (Fleiss' kappa = 0.460, Jaccard similarity = 0.464), revealing conditions where collective supervised learning performed better.

% The performance shift across studies illuminates an important point: collective learning methods excel with homogeneous preferences and larger datasets. This was evident in Study 2, where the performance of the collective MLP and CNN surpassed our approach. This finding delineates the boundary conditions for individualized versus collective preference learning strategies.

Input representation emerged as an important factor in preference learning, as demonstrated by the performance difference in Study 2 between MLP (using feature-engineered inputs) and CNN (using image inputs) models. This aligns with work on representation learning \cite{bengio2013representation} and underscores the importance of explicit feature identification, a core aspect of our dialogue-based approach, in capturing decision-making processes.

Our findings indicate that individualized language-based reward modelling is most effective under the following conditions: high preference heterogeneity, limited samples per individual, complex learning problems, and scenarios requiring feature discovery. These conditions stand in contrast to those favouring traditional reward modelling methods, which thrive with homogeneous preferences, large datasets, and simpler learning problems.

\subsection{Other Findings}
\subsubsection{Beyond Markov: Capturing Temporal Dynamics in Value Judgments}\label{section:beyond_markov}
An important finding from our study was the prevalence of non-Markovian features in participants' evaluations of respectful behaviour. This observation challenges a fundamental assumption in many reinforcement learning systems: that the current state contains all the necessary information to make a decision.
Participants often based their judgments on sequences of actions rather than single states. For instance, considering whether an agent picked up garbage before collecting an apple or whether it entered a quadrant previously visited by another agent. These non-Markovian features require information from multiple time steps, highlighting the temporal nature of many value judgments.

This finding suggests that to capture human values accurately, reward models need to consider temporal sequences and historical context, not just instantaneous states. Our approach, which takes entire trajectories as input, naturally accommodates these non-Markovian features, allowing for a more comprehensive understanding of human values.

% \subsubsection{Democratizing AI Alignment}
% Our system's ability to capture accurate reward models from participants unfamiliar with reinforcement learning suggests the potential for democratizing AI alignment. This accessibility is crucial as AI systems become increasingly integrated into daily life, suggesting that with appropriate interfaces and methodologies, individuals could shape AI behaviour according to their personal values without requiring deep technical knowledge.

\subsection{Limitations}

\subsubsection{Implementation Dependence}

Some aspects of our system are artifacts resulting from the current capabilities of large language and vision language models. For example, we textually encode trajectories so that they can be input into a language model. We chose this because vision language models performed poorly in the environments we tested. However, this may not be necessary in the future as the capabilities of LLMs and VLMs increase. That said, the core of the system, namely the overall pipeline, is not model dependent, and as capabilities increase, we expect the performance of the pipeline only to increase. 

\subsubsection{Reward Model vs. Agent Accuracy}
We focused our system and evaluation on the feedback collection and reward modelling phases of the RLHF pipeline and did not consider policy optimization (agent training). This approach aligns with recent trends in the field, where reward models themselves are the subject of evaluation \cite{lambert2024rewardbench, siththaranjan2023distributional}. We chose this focus because our core innovation is not related to policy optimization. While this leaves questions open about the behaviour that would result from training an agent with a reward model produced by our system, Kwon et al. \cite{kwon2022reward} found that RL agent accuracy mimics the reward model accuracy, suggesting that results about reward models are transferable to trained agents.

% \subsubsection*{Binary Labels}
%  Multiple participants mentioned that it was difficult to classify some trajectories as either respectful or disrespectful and that some seemed neutral. As such, the fact that our system uses binary reward could be seen as a drawback. 
%  % In RLHF it is common to use pairwise comparisons where participants would, for example, decide which of two options is ``more respectful.''
 
% \subsubsection*{Environment Complexity}
% The environment we used to evaluate our system is relatively simple and abstract. Using our system in more realistic environments could present new challenges, but it could also be easier in some ways. For example, perhaps vision language models would perform well in more realistic environments.

\subsection{Future Work}
Our research into \systemname{} opens up several promising avenues for future investigation.

\subsubsection{Interactive RLHF}
In our studies, we looked specifically into reward modelling and did not train agents with the reward models generated by our system. However, each type of reward model, including the ones generated by our system and the ones we compared to, had imperfections. These imperfections can result in imperfect agents. As such, future work can examine how to interleave reward modelling and agent training. In this way, the user can iteratively give feedback directly addressing the trained agent's imperfections. 

\subsubsection{Hybrid Approaches}
Our current work treated language-based reward modelling and supervised learning as distinct approaches. However, there's potential for developing hybrid methodologies that leverage the strengths of both. For instance, our language-based reward modelling pipeline could be used to generate additional training samples for supervised learning, similar to the approach presented by Lee et al. \cite{lee2023rlaif}.

\subsubsection{Aligning to Group vs. Individual Values}
% Our work has focused on learning reward models that mimic an individual. While this approach can be used for personalization, it can be extended to group alignment by considering the group's preferences as an aggregation of its members' individual preferences. While this method has been previously explored in the context of collective decision-making and virtual democracy \cite{noothigattu2018voting, freedman2020adapting, mohsin2021making, gordon2022jury}, applying it to align an agent's behaviour with a group's preferences is a promising avenue for future research.

While our work focused on capturing individual preferences, many real-world scenarios require alignment with collective or group values. Future research could extend our approach to group alignment by aggregating individual preferences. This direction builds on existing work in collective decision-making and virtual democracy \cite{noothigattu2018voting, freedman2020adapting, mohsin2021making, gordon2022jury}, but applies these concepts specifically to the challenge of AI alignment.
Key questions in this area include how to fairly aggregate diverse individual preferences in temporally extended scenarios, how to handle conflicts between individual and group values, and how to ensure that the resulting aligned AI systems are acceptable to all stakeholders. 
% However, this democratization raises ethical considerations. As our study showed, individual values can sometimes conflict. These conflicts highlight the need for careful consideration of how individual value alignment scales to multi-stakeholder scenarios.

% \subsection{Limitations and Future Directions}
% Our current implementation depends on the capabilities of large language and vision language models, which may evolve rapidly. Additionally, our focus on reward modelling rather than policy optimization leaves open questions about how our approach translates to actual agent behaviour.

% These limitations point to directions for future research. Exploring interactive RLHF, interleaving reward modeling and agent training could provide a more holistic approach to AI alignment. Extending our approach to group value alignment, building on work in collective decision-making and virtual democracy \cite{noothigattu2018voting, freedman2020adapting, mohsin2021making, gordon2022jury}, could broaden the applicability of our approach.
\section{Conclusion}

We developed a system, \systemname{}, that aids users with no particular experience in machine learning in designing a reward model that can be used to train agents in alignment with their individual understanding of values. Our findings demonstrate that \systemname{} effectively captures diverse and subjective value definitions through an interactive, reflective, language-based process. By enabling users to engage deeply with the nuances of their values, the system provides an individual-conscious approach to AI alignment. The studies on ``respect'' and ethical decision-making in autonomous vehicles illustrate the system's capability to accommodate a wide range of value-aligned behaviors. This method can be used to personalizes AI agents and as the foundation for interpretable and representative collective alignment strategies

% \input{content/8_appendix}

% \begin{acks}
% \end{acks}

\bibliographystyle{ACM-Reference-Format}
\bibliography{main}


\begin{thebibliography}{78}

%%% the Web address.

\ifx \showCODEN    \undefined \def \showCODEN     #1{\unskip}     \fi
\ifx \showDOI      \undefined \def \showDOI       #1{#1}\fi
\ifx \showISBNx    \undefined \def \showISBNx     #1{\unskip}     \fi
\ifx \showISBNxiii \undefined \def \showISBNxiii  #1{\unskip}     \fi
\ifx \showISSN     \undefined \def \showISSN      #1{\unskip}     \fi
\ifx \showLCCN     \undefined \def \showLCCN      #1{\unskip}     \fi
\ifx \shownote     \undefined \def \shownote      #1{#1}          \fi
\ifx \showarticletitle \undefined \def \showarticletitle #1{#1}   \fi
\ifx \showURL      \undefined \def \showURL       {\relax}        \fi
% invisible to TeX
\providecommand\bibfield[2]{#2}
\providecommand\bibinfo[2]{#2}
\providecommand\natexlab[1]{#1}
\providecommand\showeprint[2][]{arXiv:#2}

\bibitem[Abbeel and Ng(2004)]%
        {abbeel2004apprenticeship}
\bibfield{author}{\bibinfo{person}{Pieter Abbeel} {and} \bibinfo{person}{Andrew~Y Ng}.} \bibinfo{year}{2004}\natexlab{}.
\newblock \showarticletitle{Apprenticeship learning via inverse reinforcement learning}. In \bibinfo{booktitle}{\emph{Proceedings of the twenty-first international conference on Machine learning}}. \bibinfo{pages}{1}.
\newblock


\bibitem[Adomavicius and Tuzhilin(2005)]%
        {adomavicius2005toward}
\bibfield{author}{\bibinfo{person}{Gediminas Adomavicius} {and} \bibinfo{person}{Alexander Tuzhilin}.} \bibinfo{year}{2005}\natexlab{}.
\newblock \showarticletitle{Toward the next generation of recommender systems: A survey of the state-of-the-art and possible extensions}.
\newblock \bibinfo{journal}{\emph{IEEE transactions on knowledge and data engineering}} \bibinfo{volume}{17}, \bibinfo{number}{6} (\bibinfo{year}{2005}), \bibinfo{pages}{734--749}.
\newblock


\bibitem[Agapiou et~al\mbox{.}(2022)]%
        {agapiou2022melting}
\bibfield{author}{\bibinfo{person}{John~P Agapiou}, \bibinfo{person}{Alexander~Sasha Vezhnevets}, \bibinfo{person}{Edgar~A Du{\'e}{\~n}ez-Guzm{\'a}n}, \bibinfo{person}{Jayd Matyas}, \bibinfo{person}{Yiran Mao}, \bibinfo{person}{Peter Sunehag}, \bibinfo{person}{Raphael K{\"o}ster}, \bibinfo{person}{Udari Madhushani}, \bibinfo{person}{Kavya Kopparapu}, \bibinfo{person}{Ramona Comanescu}, {et~al\mbox{.}}} \bibinfo{year}{2022}\natexlab{}.
\newblock \showarticletitle{Melting Pot 2.0}.
\newblock \bibinfo{journal}{\emph{arXiv preprint arXiv:2211.13746}} (\bibinfo{year}{2022}).
\newblock


\bibitem[Amodei et~al\mbox{.}(2016)]%
        {amodei_concrete_2016}
\bibfield{author}{\bibinfo{person}{Dario Amodei}, \bibinfo{person}{Chris Olah}, \bibinfo{person}{Jacob Steinhardt}, \bibinfo{person}{Paul Christiano}, \bibinfo{person}{John Schulman}, {and} \bibinfo{person}{Dan Mané}.} \bibinfo{year}{2016}\natexlab{}.
\newblock \bibinfo{title}{Concrete {Problems} in {AI} {Safety}}.
\newblock
\newblock
\urldef\tempurl%
\url{https://doi.org/10.48550/arXiv.1606.06565}
\showDOI{\tempurl}
\newblock
\shownote{arXiv:1606.06565 [cs]}.


\bibitem[Awad et~al\mbox{.}(2018)]%
        {awad2018moral}
\bibfield{author}{\bibinfo{person}{Edmond Awad}, \bibinfo{person}{Sohan Dsouza}, \bibinfo{person}{Richard Kim}, \bibinfo{person}{Jonathan Schulz}, \bibinfo{person}{Joseph Henrich}, \bibinfo{person}{Azim Shariff}, \bibinfo{person}{Jean-Fran{\c{c}}ois Bonnefon}, {and} \bibinfo{person}{Iyad Rahwan}.} \bibinfo{year}{2018}\natexlab{}.
\newblock \showarticletitle{The moral machine experiment}.
\newblock \bibinfo{journal}{\emph{Nature}} \bibinfo{volume}{563}, \bibinfo{number}{7729} (\bibinfo{year}{2018}), \bibinfo{pages}{59--64}.
\newblock


\bibitem[Bai et~al\mbox{.}(2022a)]%
        {bai_training_2022}
\bibfield{author}{\bibinfo{person}{Yuntao Bai}, \bibinfo{person}{Andy Jones}, \bibinfo{person}{Kamal Ndousse}, \bibinfo{person}{Amanda Askell}, \bibinfo{person}{Anna Chen}, \bibinfo{person}{Nova DasSarma}, \bibinfo{person}{Dawn Drain}, \bibinfo{person}{Stanislav Fort}, \bibinfo{person}{Deep Ganguli}, \bibinfo{person}{Tom Henighan}, \bibinfo{person}{Nicholas Joseph}, \bibinfo{person}{Saurav Kadavath}, \bibinfo{person}{Jackson Kernion}, \bibinfo{person}{Tom Conerly}, \bibinfo{person}{Sheer El-Showk}, \bibinfo{person}{Nelson Elhage}, \bibinfo{person}{Zac Hatfield-Dodds}, \bibinfo{person}{Danny Hernandez}, \bibinfo{person}{Tristan Hume}, \bibinfo{person}{Scott Johnston}, \bibinfo{person}{Shauna Kravec}, \bibinfo{person}{Liane Lovitt}, \bibinfo{person}{Neel Nanda}, \bibinfo{person}{Catherine Olsson}, \bibinfo{person}{Dario Amodei}, \bibinfo{person}{Tom Brown}, \bibinfo{person}{Jack Clark}, \bibinfo{person}{Sam McCandlish}, \bibinfo{person}{Chris Olah}, \bibinfo{person}{Ben Mann}, {and} \bibinfo{person}{Jared
  Kaplan}.} \bibinfo{year}{2022}\natexlab{a}.
\newblock \bibinfo{title}{Training a {Helpful} and {Harmless} {Assistant} with {Reinforcement} {Learning} from {Human} {Feedback}}.
\newblock
\newblock
\urldef\tempurl%
\url{https://doi.org/10.48550/arXiv.2204.05862}
\showDOI{\tempurl}
\newblock
\shownote{arXiv:2204.05862 [cs]}.


\bibitem[Bai et~al\mbox{.}(2022b)]%
        {bai_constitutional_2022}
\bibfield{author}{\bibinfo{person}{Yuntao Bai}, \bibinfo{person}{Saurav Kadavath}, \bibinfo{person}{Sandipan Kundu}, \bibinfo{person}{Amanda Askell}, \bibinfo{person}{Jackson Kernion}, \bibinfo{person}{Andy Jones}, \bibinfo{person}{Anna Chen}, \bibinfo{person}{Anna Goldie}, \bibinfo{person}{Azalia Mirhoseini}, \bibinfo{person}{Cameron McKinnon}, \bibinfo{person}{Carol Chen}, \bibinfo{person}{Catherine Olsson}, \bibinfo{person}{Christopher Olah}, \bibinfo{person}{Danny Hernandez}, \bibinfo{person}{Dawn Drain}, \bibinfo{person}{Deep Ganguli}, \bibinfo{person}{Dustin Li}, \bibinfo{person}{Eli Tran-Johnson}, \bibinfo{person}{Ethan Perez}, \bibinfo{person}{Jamie Kerr}, \bibinfo{person}{Jared Mueller}, \bibinfo{person}{Jeffrey Ladish}, \bibinfo{person}{Joshua Landau}, \bibinfo{person}{Kamal Ndousse}, \bibinfo{person}{Kamile Lukosuite}, \bibinfo{person}{Liane Lovitt}, \bibinfo{person}{Michael Sellitto}, \bibinfo{person}{Nelson Elhage}, \bibinfo{person}{Nicholas Schiefer}, \bibinfo{person}{Noemi Mercado},
  \bibinfo{person}{Nova DasSarma}, \bibinfo{person}{Robert Lasenby}, \bibinfo{person}{Robin Larson}, \bibinfo{person}{Sam Ringer}, \bibinfo{person}{Scott Johnston}, \bibinfo{person}{Shauna Kravec}, \bibinfo{person}{Sheer~El Showk}, \bibinfo{person}{Stanislav Fort}, \bibinfo{person}{Tamera Lanham}, \bibinfo{person}{Timothy Telleen-Lawton}, \bibinfo{person}{Tom Conerly}, \bibinfo{person}{Tom Henighan}, \bibinfo{person}{Tristan Hume}, \bibinfo{person}{Samuel~R. Bowman}, \bibinfo{person}{Zac Hatfield-Dodds}, \bibinfo{person}{Ben Mann}, \bibinfo{person}{Dario Amodei}, \bibinfo{person}{Nicholas Joseph}, \bibinfo{person}{Sam McCandlish}, \bibinfo{person}{Tom Brown}, {and} \bibinfo{person}{Jared Kaplan}.} \bibinfo{year}{2022}\natexlab{b}.
\newblock \bibinfo{title}{Constitutional {AI}: {Harmlessness} from {AI} {Feedback}}.
\newblock
\newblock
\urldef\tempurl%
\url{https://arxiv.org/abs/2212.08073v1}
\showURL{%
\tempurl}


\bibitem[Bengio et~al\mbox{.}(2013)]%
        {bengio2013representation}
\bibfield{author}{\bibinfo{person}{Yoshua Bengio}, \bibinfo{person}{Aaron Courville}, {and} \bibinfo{person}{Pascal Vincent}.} \bibinfo{year}{2013}\natexlab{}.
\newblock \showarticletitle{Representation learning: A review and new perspectives}.
\newblock \bibinfo{journal}{\emph{IEEE transactions on pattern analysis and machine intelligence}} \bibinfo{volume}{35}, \bibinfo{number}{8} (\bibinfo{year}{2013}), \bibinfo{pages}{1798--1828}.
\newblock


\bibitem[Bentvelzen et~al\mbox{.}(2022)]%
        {bentvelzen_revisiting_2022}
\bibfield{author}{\bibinfo{person}{Marit Bentvelzen}, \bibinfo{person}{Paweł~W. Woźniak}, \bibinfo{person}{Pia~S.F. Herbes}, \bibinfo{person}{Evropi Stefanidi}, {and} \bibinfo{person}{Jasmin Niess}.} \bibinfo{year}{2022}\natexlab{}.
\newblock \showarticletitle{Revisiting {Reflection} in {HCI}: {Four} {Design} {Resources} for {Technologies} that {Support} {Reflection}}.
\newblock \bibinfo{journal}{\emph{Proceedings of the ACM on Interactive, Mobile, Wearable and Ubiquitous Technologies}} \bibinfo{volume}{6}, \bibinfo{number}{1} (\bibinfo{date}{March} \bibinfo{year}{2022}), \bibinfo{pages}{2:1--2:27}.
\newblock
\urldef\tempurl%
\url{https://doi.org/10.1145/3517233}
\showDOI{\tempurl}


\bibitem[Berlin(1969)]%
        {berlin1969four}
\bibfield{author}{\bibinfo{person}{Isaiah Berlin}.} \bibinfo{year}{1969}\natexlab{}.
\newblock \bibinfo{title}{Four essays on liberty}.
\newblock
\newblock


\bibitem[Biyik et~al\mbox{.}(2023)]%
        {biyik2023preference}
\bibfield{author}{\bibinfo{person}{Erdem Biyik}, \bibinfo{person}{Fan Yao}, \bibinfo{person}{Yinlam Chow}, \bibinfo{person}{Alex Haig}, \bibinfo{person}{Chih-wei Hsu}, \bibinfo{person}{Mohammad Ghavamzadeh}, {and} \bibinfo{person}{Craig Boutilier}.} \bibinfo{year}{2023}\natexlab{}.
\newblock \showarticletitle{Preference Elicitation with Soft Attributes in Interactive Recommendation}.
\newblock \bibinfo{journal}{\emph{arXiv preprint arXiv:2311.02085}} (\bibinfo{year}{2023}).
\newblock


\bibitem[Braun and Clarke(2006)]%
        {braun2006using}
\bibfield{author}{\bibinfo{person}{Virginia Braun} {and} \bibinfo{person}{Victoria Clarke}.} \bibinfo{year}{2006}\natexlab{}.
\newblock \showarticletitle{Using thematic analysis in psychology}.
\newblock \bibinfo{journal}{\emph{Qualitative research in psychology}} \bibinfo{volume}{3}, \bibinfo{number}{2} (\bibinfo{year}{2006}), \bibinfo{pages}{77--101}.
\newblock


\bibitem[Bridge and Sawilowsky(1999)]%
        {bridge1999increasing}
\bibfield{author}{\bibinfo{person}{Patrick~D Bridge} {and} \bibinfo{person}{Shlomo~S Sawilowsky}.} \bibinfo{year}{1999}\natexlab{}.
\newblock \showarticletitle{Increasing physicians’ awareness of the impact of statistics on research outcomes: comparative power of the t-test and Wilcoxon rank-sum test in small samples applied research}.
\newblock \bibinfo{journal}{\emph{Journal of clinical epidemiology}} \bibinfo{volume}{52}, \bibinfo{number}{3} (\bibinfo{year}{1999}), \bibinfo{pages}{229--235}.
\newblock


\bibitem[Brown et~al\mbox{.}(2020)]%
        {brown_language_2020}
\bibfield{author}{\bibinfo{person}{Tom~B. Brown}, \bibinfo{person}{Benjamin Mann}, \bibinfo{person}{Nick Ryder}, \bibinfo{person}{Melanie Subbiah}, \bibinfo{person}{Jared Kaplan}, \bibinfo{person}{Prafulla Dhariwal}, \bibinfo{person}{Arvind Neelakantan}, \bibinfo{person}{Pranav Shyam}, \bibinfo{person}{Girish Sastry}, \bibinfo{person}{Amanda Askell}, \bibinfo{person}{Sandhini Agarwal}, \bibinfo{person}{Ariel Herbert-Voss}, \bibinfo{person}{Gretchen Krueger}, \bibinfo{person}{Tom Henighan}, \bibinfo{person}{Rewon Child}, \bibinfo{person}{Aditya Ramesh}, \bibinfo{person}{Daniel~M. Ziegler}, \bibinfo{person}{Jeffrey Wu}, \bibinfo{person}{Clemens Winter}, \bibinfo{person}{Christopher Hesse}, \bibinfo{person}{Mark Chen}, \bibinfo{person}{Eric Sigler}, \bibinfo{person}{Mateusz Litwin}, \bibinfo{person}{Scott Gray}, \bibinfo{person}{Benjamin Chess}, \bibinfo{person}{Jack Clark}, \bibinfo{person}{Christopher Berner}, \bibinfo{person}{Sam McCandlish}, \bibinfo{person}{Alec Radford}, \bibinfo{person}{Ilya Sutskever},
  {and} \bibinfo{person}{Dario Amodei}.} \bibinfo{year}{2020}\natexlab{}.
\newblock \bibinfo{title}{Language {Models} are {Few}-{Shot} {Learners}}.
\newblock
\newblock
\urldef\tempurl%
\url{https://doi.org/10.48550/arXiv.2005.14165}
\showDOI{\tempurl}
\newblock
\shownote{arXiv:2005.14165 [cs]}.


\bibitem[Casper et~al\mbox{.}(2023)]%
        {casper_open_nodate}
\bibfield{author}{\bibinfo{person}{Stephen Casper}, \bibinfo{person}{Xander Davies}, \bibinfo{person}{Claudia Shi}, \bibinfo{person}{Thomas~Krendl Gilbert}, \bibinfo{person}{J{\'e}r{\'e}my Scheurer}, \bibinfo{person}{Javier Rando}, \bibinfo{person}{Rachel Freedman}, \bibinfo{person}{Tomasz Korbak}, \bibinfo{person}{David Lindner}, \bibinfo{person}{Pedro Freire}, {et~al\mbox{.}}} \bibinfo{year}{2023}\natexlab{}.
\newblock \showarticletitle{Open problems and fundamental limitations of reinforcement learning from human feedback}.
\newblock \bibinfo{journal}{\emph{arXiv preprint arXiv:2307.15217}} (\bibinfo{year}{2023}).
\newblock


\bibitem[Cecchini(2021)]%
        {cecchini2021dual}
\bibfield{author}{\bibinfo{person}{Dario Cecchini}.} \bibinfo{year}{2021}\natexlab{}.
\newblock \showarticletitle{Dual-process reflective equilibrium: rethinking the interplay between intuition and reflection in moral reasoning}.
\newblock \bibinfo{journal}{\emph{Philosophical Explorations}} \bibinfo{volume}{24}, \bibinfo{number}{3} (\bibinfo{year}{2021}), \bibinfo{pages}{295--311}.
\newblock


\bibitem[Christakopoulou et~al\mbox{.}(2016)]%
        {christakopoulou2016towards}
\bibfield{author}{\bibinfo{person}{Konstantina Christakopoulou}, \bibinfo{person}{Filip Radlinski}, {and} \bibinfo{person}{Katja Hofmann}.} \bibinfo{year}{2016}\natexlab{}.
\newblock \showarticletitle{Towards conversational recommender systems}. In \bibinfo{booktitle}{\emph{Proceedings of the 22nd ACM SIGKDD international conference on knowledge discovery and data mining}}. \bibinfo{pages}{815--824}.
\newblock


\bibitem[Christiano et~al\mbox{.}(2017)]%
        {christiano_deep_2017}
\bibfield{author}{\bibinfo{person}{Paul Christiano}, \bibinfo{person}{Jan Leike}, \bibinfo{person}{Tom~B. Brown}, \bibinfo{person}{Miljan Martic}, \bibinfo{person}{Shane Legg}, {and} \bibinfo{person}{Dario Amodei}.} \bibinfo{year}{2017}\natexlab{}.
\newblock \bibinfo{title}{Deep reinforcement learning from human preferences}.
\newblock
\newblock
\urldef\tempurl%
\url{https://arxiv.org/abs/1706.03741v4}
\showURL{%
\tempurl}


\bibitem[Costa~Filho et~al\mbox{.}(2019)]%
        {costa2019effects}
\bibfield{author}{\bibinfo{person}{Galileu~B Costa~Filho}, \bibinfo{person}{Alexandre~S Moura}, \bibinfo{person}{Paulo~R Brand{\~a}o}, \bibinfo{person}{Henk~G Schmidt}, {and} \bibinfo{person}{Silvia Mamede}.} \bibinfo{year}{2019}\natexlab{}.
\newblock \showarticletitle{Effects of deliberate reflection on diagnostic accuracy, confidence and diagnostic calibration in dermatology}.
\newblock \bibinfo{journal}{\emph{Perspectives on Medical Education}}  \bibinfo{volume}{8} (\bibinfo{year}{2019}), \bibinfo{pages}{230--236}.
\newblock


\bibitem[Dewey(2014)]%
        {dewey2014reinforcement}
\bibfield{author}{\bibinfo{person}{Daniel Dewey}.} \bibinfo{year}{2014}\natexlab{}.
\newblock \showarticletitle{Reinforcement learning and the reward engineering principle}. In \bibinfo{booktitle}{\emph{2014 AAAI Spring Symposium Series}}.
\newblock


\bibitem[Evans(2019)]%
        {evans2019reflections}
\bibfield{author}{\bibinfo{person}{Jonathan St~BT Evans}.} \bibinfo{year}{2019}\natexlab{}.
\newblock \showarticletitle{Reflections on reflection: The nature and function of type 2 processes in dual-process theories of reasoning}.
\newblock \bibinfo{journal}{\emph{Thinking \& Reasoning}} \bibinfo{volume}{25}, \bibinfo{number}{4} (\bibinfo{year}{2019}), \bibinfo{pages}{383--415}.
\newblock


\bibitem[Everitt and Hutter(2016)]%
        {everitt2016avoiding}
\bibfield{author}{\bibinfo{person}{Tom Everitt} {and} \bibinfo{person}{Marcus Hutter}.} \bibinfo{year}{2016}\natexlab{}.
\newblock \showarticletitle{Avoiding wireheading with value reinforcement learning}. In \bibinfo{booktitle}{\emph{Artificial General Intelligence: 9th International Conference, AGI 2016, New York, NY, USA, July 16-19, 2016, Proceedings 9}}. Springer, \bibinfo{pages}{12--22}.
\newblock


\bibitem[Fernandes et~al\mbox{.}(2021)]%
        {fernandes2021adding}
\bibfield{author}{\bibinfo{person}{Rachel Aparecida~Ferreira Fernandes}, \bibinfo{person}{Leandro~Fernandes Malloy-Diniz}, \bibinfo{person}{Marcos~Carvalho de Vasconcellos}, \bibinfo{person}{Paulo Augusto~Moreira Camargos}, {and} \bibinfo{person}{C{\'a}ssio Ibiapina}.} \bibinfo{year}{2021}\natexlab{}.
\newblock \showarticletitle{Adding guidance to deliberate reflection improves medical student’s diagnostic accuracy}.
\newblock \bibinfo{journal}{\emph{Medical Education}} \bibinfo{volume}{55}, \bibinfo{number}{10} (\bibinfo{year}{2021}), \bibinfo{pages}{1161--1171}.
\newblock


\bibitem[Fessl et~al\mbox{.}(2017)]%
        {fessl2017known}
\bibfield{author}{\bibinfo{person}{Angela Fessl}, \bibinfo{person}{Oliver Blunk}, \bibinfo{person}{Michael Prilla}, {and} \bibinfo{person}{Viktoria Pammer}.} \bibinfo{year}{2017}\natexlab{}.
\newblock \showarticletitle{The known universe of reflection guidance: a literature review}.
\newblock \bibinfo{journal}{\emph{International journal of technology enhanced learning}} \bibinfo{volume}{9}, \bibinfo{number}{2-3} (\bibinfo{year}{2017}), \bibinfo{pages}{103--125}.
\newblock


\bibitem[Fleck and Fitzpatrick(2010)]%
        {fleck2010reflecting}
\bibfield{author}{\bibinfo{person}{Rowanne Fleck} {and} \bibinfo{person}{Geraldine Fitzpatrick}.} \bibinfo{year}{2010}\natexlab{}.
\newblock \showarticletitle{Reflecting on reflection: framing a design landscape}. In \bibinfo{booktitle}{\emph{Proceedings of the 22nd conference of the computer-human interaction special interest group of australia on computer-human interaction}}. \bibinfo{pages}{216--223}.
\newblock


\bibitem[Freedman et~al\mbox{.}(2020)]%
        {freedman2020adapting}
\bibfield{author}{\bibinfo{person}{Rachel Freedman}, \bibinfo{person}{Jana~Schaich Borg}, \bibinfo{person}{Walter Sinnott-Armstrong}, \bibinfo{person}{John~P Dickerson}, {and} \bibinfo{person}{Vincent Conitzer}.} \bibinfo{year}{2020}\natexlab{}.
\newblock \showarticletitle{Adapting a kidney exchange algorithm to align with human values}.
\newblock \bibinfo{journal}{\emph{Artificial Intelligence}}  \bibinfo{volume}{283} (\bibinfo{year}{2020}), \bibinfo{pages}{103261}.
\newblock


\bibitem[Friedman et~al\mbox{.}(2013)]%
        {friedman2013value}
\bibfield{author}{\bibinfo{person}{Batya Friedman}, \bibinfo{person}{Peter~H Kahn}, \bibinfo{person}{Alan Borning}, {and} \bibinfo{person}{Alina Huldtgren}.} \bibinfo{year}{2013}\natexlab{}.
\newblock \showarticletitle{Value sensitive design and information systems}.
\newblock \bibinfo{journal}{\emph{Early engagement and new technologies: Opening up the laboratory}} (\bibinfo{year}{2013}), \bibinfo{pages}{55--95}.
\newblock


\bibitem[Ghajargar et~al\mbox{.}(2018)]%
        {ghajargar2018designing}
\bibfield{author}{\bibinfo{person}{Maliheh Ghajargar}, \bibinfo{person}{Mikael Wiberg}, {and} \bibinfo{person}{Erik Stolterman}.} \bibinfo{year}{2018}\natexlab{}.
\newblock \showarticletitle{Designing IoT systems that support reflective thinking: A relational approach}.
\newblock \bibinfo{journal}{\emph{International Journal of Design}} \bibinfo{volume}{12}, \bibinfo{number}{1} (\bibinfo{year}{2018}), \bibinfo{pages}{21--35}.
\newblock


\bibitem[Gordon et~al\mbox{.}(2022)]%
        {gordon2022jury}
\bibfield{author}{\bibinfo{person}{Mitchell~L Gordon}, \bibinfo{person}{Michelle~S Lam}, \bibinfo{person}{Joon~Sung Park}, \bibinfo{person}{Kayur Patel}, \bibinfo{person}{Jeff Hancock}, \bibinfo{person}{Tatsunori Hashimoto}, {and} \bibinfo{person}{Michael~S Bernstein}.} \bibinfo{year}{2022}\natexlab{}.
\newblock \showarticletitle{Jury learning: Integrating dissenting voices into machine learning models}. In \bibinfo{booktitle}{\emph{Proceedings of the 2022 CHI Conference on Human Factors in Computing Systems}}. \bibinfo{pages}{1--19}.
\newblock


\bibitem[Graus and Willemsen(2015)]%
        {graus2015improving}
\bibfield{author}{\bibinfo{person}{Mark~P Graus} {and} \bibinfo{person}{Martijn~C Willemsen}.} \bibinfo{year}{2015}\natexlab{}.
\newblock \showarticletitle{Improving the user experience during cold start through choice-based preference elicitation}. In \bibinfo{booktitle}{\emph{Proceedings of the 9th ACM Conference on Recommender Systems}}. \bibinfo{pages}{273--276}.
\newblock


\bibitem[Hauser et~al\mbox{.}(2014)]%
        {hauser2014self}
\bibfield{author}{\bibinfo{person}{John~R Hauser}, \bibinfo{person}{Songting Dong}, {and} \bibinfo{person}{Min Ding}.} \bibinfo{year}{2014}\natexlab{}.
\newblock \showarticletitle{Self-reflection and articulated consumer preferences}.
\newblock \bibinfo{journal}{\emph{Journal of Product Innovation Management}} \bibinfo{volume}{31}, \bibinfo{number}{1} (\bibinfo{year}{2014}), \bibinfo{pages}{17--32}.
\newblock


\bibitem[Ifenthaler(2012)]%
        {ifenthaler2012determining}
\bibfield{author}{\bibinfo{person}{Dirk Ifenthaler}.} \bibinfo{year}{2012}\natexlab{}.
\newblock \showarticletitle{Determining the effectiveness of prompts for self-regulated learning in problem-solving scenarios}.
\newblock \bibinfo{journal}{\emph{Journal of Educational Technology \& Society}} \bibinfo{volume}{15}, \bibinfo{number}{1} (\bibinfo{year}{2012}), \bibinfo{pages}{38--52}.
\newblock


\bibitem[Jaccard(1912)]%
        {jaccard1912distribution}
\bibfield{author}{\bibinfo{person}{Paul Jaccard}.} \bibinfo{year}{1912}\natexlab{}.
\newblock \showarticletitle{The distribution of the flora in the alpine zone. 1}.
\newblock \bibinfo{journal}{\emph{New phytologist}} \bibinfo{volume}{11}, \bibinfo{number}{2} (\bibinfo{year}{1912}), \bibinfo{pages}{37--50}.
\newblock


\bibitem[Kocielnik et~al\mbox{.}(2018a)]%
        {kocielnik_reflection_2018}
\bibfield{author}{\bibinfo{person}{Rafal Kocielnik}, \bibinfo{person}{Lillian Xiao}, \bibinfo{person}{Daniel Avrahami}, {and} \bibinfo{person}{Gary Hsieh}.} \bibinfo{year}{2018}\natexlab{a}.
\newblock \showarticletitle{Reflection {Companion}: {A} {Conversational} {System} for {Engaging} {Users} in {Reflection} on {Physical} {Activity}}.
\newblock \bibinfo{journal}{\emph{Proceedings of the ACM on Interactive, Mobile, Wearable and Ubiquitous Technologies}} \bibinfo{volume}{2}, \bibinfo{number}{2} (\bibinfo{date}{July} \bibinfo{year}{2018}), \bibinfo{pages}{70:1--70:26}.
\newblock
\urldef\tempurl%
\url{https://doi.org/10.1145/3214273}
\showDOI{\tempurl}


\bibitem[Kocielnik et~al\mbox{.}(2018b)]%
        {kocielnik2018reflection}
\bibfield{author}{\bibinfo{person}{Rafal Kocielnik}, \bibinfo{person}{Lillian Xiao}, \bibinfo{person}{Daniel Avrahami}, {and} \bibinfo{person}{Gary Hsieh}.} \bibinfo{year}{2018}\natexlab{b}.
\newblock \showarticletitle{Reflection companion: a conversational system for engaging users in reflection on physical activity}.
\newblock \bibinfo{journal}{\emph{Proceedings of the ACM on Interactive, Mobile, Wearable and Ubiquitous Technologies}} \bibinfo{volume}{2}, \bibinfo{number}{2} (\bibinfo{year}{2018}), \bibinfo{pages}{1--26}.
\newblock


\bibitem[Kojima et~al\mbox{.}(2022)]%
        {kojima2022large}
\bibfield{author}{\bibinfo{person}{Takeshi Kojima}, \bibinfo{person}{Shixiang~Shane Gu}, \bibinfo{person}{Machel Reid}, \bibinfo{person}{Yutaka Matsuo}, {and} \bibinfo{person}{Yusuke Iwasawa}.} \bibinfo{year}{2022}\natexlab{}.
\newblock \showarticletitle{Large language models are zero-shot reasoners}.
\newblock \bibinfo{journal}{\emph{Advances in neural information processing systems}}  \bibinfo{volume}{35} (\bibinfo{year}{2022}), \bibinfo{pages}{22199--22213}.
\newblock


\bibitem[Kwon et~al\mbox{.}(2022)]%
        {kwon2022reward}
\bibfield{author}{\bibinfo{person}{Minae Kwon}, \bibinfo{person}{Sang~Michael Xie}, \bibinfo{person}{Kalesha Bullard}, {and} \bibinfo{person}{Dorsa Sadigh}.} \bibinfo{year}{2022}\natexlab{}.
\newblock \showarticletitle{Reward Design with Language Models}. In \bibinfo{booktitle}{\emph{The Eleventh International Conference on Learning Representations}}.
\newblock


\bibitem[Lambert et~al\mbox{.}(2024)]%
        {lambert2024rewardbench}
\bibfield{author}{\bibinfo{person}{Nathan Lambert}, \bibinfo{person}{Valentina Pyatkin}, \bibinfo{person}{Jacob Morrison}, \bibinfo{person}{LJ Miranda}, \bibinfo{person}{Bill~Yuchen Lin}, \bibinfo{person}{Khyathi Chandu}, \bibinfo{person}{Nouha Dziri}, \bibinfo{person}{Sachin Kumar}, \bibinfo{person}{Tom Zick}, \bibinfo{person}{Yejin Choi}, {et~al\mbox{.}}} \bibinfo{year}{2024}\natexlab{}.
\newblock \showarticletitle{Rewardbench: Evaluating reward models for language modeling}.
\newblock \bibinfo{journal}{\emph{arXiv preprint arXiv:2403.13787}} (\bibinfo{year}{2024}).
\newblock


\bibitem[Landis and Koch(1977)]%
        {landis1977measurement}
\bibfield{author}{\bibinfo{person}{J~Richard Landis} {and} \bibinfo{person}{Gary~G Koch}.} \bibinfo{year}{1977}\natexlab{}.
\newblock \showarticletitle{The measurement of observer agreement for categorical data}.
\newblock \bibinfo{journal}{\emph{biometrics}} (\bibinfo{year}{1977}), \bibinfo{pages}{159--174}.
\newblock


\bibitem[Le~Dantec et~al\mbox{.}(2009)]%
        {le2009values}
\bibfield{author}{\bibinfo{person}{Christopher~A Le~Dantec}, \bibinfo{person}{Erika~Shehan Poole}, {and} \bibinfo{person}{Susan~P Wyche}.} \bibinfo{year}{2009}\natexlab{}.
\newblock \showarticletitle{Values as lived experience: evolving value sensitive design in support of value discovery}. In \bibinfo{booktitle}{\emph{Proceedings of the SIGCHI conference on human factors in computing systems}}. \bibinfo{pages}{1141--1150}.
\newblock


\bibitem[Lee et~al\mbox{.}(2023)]%
        {lee2023rlaif}
\bibfield{author}{\bibinfo{person}{Harrison Lee}, \bibinfo{person}{Samrat Phatale}, \bibinfo{person}{Hassan Mansoor}, \bibinfo{person}{Thomas Mesnard}, \bibinfo{person}{Johan Ferret}, \bibinfo{person}{Kellie Lu}, \bibinfo{person}{Colton Bishop}, \bibinfo{person}{Ethan Hall}, \bibinfo{person}{Victor Carbune}, \bibinfo{person}{Abhinav Rastogi}, {et~al\mbox{.}}} \bibinfo{year}{2023}\natexlab{}.
\newblock \showarticletitle{Rlaif: Scaling reinforcement learning from human feedback with ai feedback}.
\newblock \bibinfo{journal}{\emph{arXiv preprint arXiv:2309.00267}} (\bibinfo{year}{2023}).
\newblock


\bibitem[Lindstr{\"o}m et~al\mbox{.}(2006)]%
        {lindstrom2006affective}
\bibfield{author}{\bibinfo{person}{Madelene Lindstr{\"o}m}, \bibinfo{person}{Anna St{\aa}hl}, \bibinfo{person}{Kristina H{\"o}{\"o}k}, \bibinfo{person}{Petra Sundstr{\"o}m}, \bibinfo{person}{Jarmo Laaksolathi}, \bibinfo{person}{Marco Combetto}, \bibinfo{person}{Alex Taylor}, {and} \bibinfo{person}{Roberto Bresin}.} \bibinfo{year}{2006}\natexlab{}.
\newblock \showarticletitle{Affective diary: designing for bodily expressiveness and self-reflection}. In \bibinfo{booktitle}{\emph{CHI'06 extended abstracts on Human factors in computing systems}}. \bibinfo{pages}{1037--1042}.
\newblock


\bibitem[Llorente and Guerrero(2011)]%
        {llorente2011increasing}
\bibfield{author}{\bibinfo{person}{Maria~Salam{\'o} Llorente} {and} \bibinfo{person}{Sergio~Escalera Guerrero}.} \bibinfo{year}{2011}\natexlab{}.
\newblock \showarticletitle{Increasing retrieval quality in conversational recommenders}.
\newblock \bibinfo{journal}{\emph{IEEE Transactions on Knowledge and Data Engineering}} \bibinfo{volume}{24}, \bibinfo{number}{10} (\bibinfo{year}{2011}), \bibinfo{pages}{1876--1888}.
\newblock


\bibitem[Loepp et~al\mbox{.}(2014)]%
        {loepp2014choice}
\bibfield{author}{\bibinfo{person}{Benedikt Loepp}, \bibinfo{person}{Tim Hussein}, {and} \bibinfo{person}{J{\"u}ergen Ziegler}.} \bibinfo{year}{2014}\natexlab{}.
\newblock \showarticletitle{Choice-based preference elicitation for collaborative filtering recommender systems}. In \bibinfo{booktitle}{\emph{Proceedings of the SIGCHI Conference on Human Factors in Computing Systems}}. \bibinfo{pages}{3085--3094}.
\newblock


\bibitem[Mamede et~al\mbox{.}(2008)]%
        {mamede2008effects}
\bibfield{author}{\bibinfo{person}{Silvia Mamede}, \bibinfo{person}{Henk~G Schmidt}, {and} \bibinfo{person}{J{\'u}lio~C{\'e}sar Penaforte}.} \bibinfo{year}{2008}\natexlab{}.
\newblock \showarticletitle{Effects of reflective practice on the accuracy of medical diagnoses}.
\newblock \bibinfo{journal}{\emph{Medical education}} \bibinfo{volume}{42}, \bibinfo{number}{5} (\bibinfo{year}{2008}), \bibinfo{pages}{468--475}.
\newblock


\bibitem[Metelli et~al\mbox{.}(2023)]%
        {metelli2023towards}
\bibfield{author}{\bibinfo{person}{Alberto~Maria Metelli}, \bibinfo{person}{Filippo Lazzati}, {and} \bibinfo{person}{Marcello Restelli}.} \bibinfo{year}{2023}\natexlab{}.
\newblock \showarticletitle{Towards theoretical understanding of inverse reinforcement learning}. In \bibinfo{booktitle}{\emph{International Conference on Machine Learning}}. PMLR, \bibinfo{pages}{24555--24591}.
\newblock


\bibitem[Mohsin et~al\mbox{.}(2021)]%
        {mohsin2021making}
\bibfield{author}{\bibinfo{person}{Farhad Mohsin}, \bibinfo{person}{Lei Luo}, \bibinfo{person}{Wufei Ma}, \bibinfo{person}{Inwon Kang}, \bibinfo{person}{Zhibing Zhao}, \bibinfo{person}{Ao Liu}, \bibinfo{person}{Rohit Vaish}, {and} \bibinfo{person}{Lirong Xia}.} \bibinfo{year}{2021}\natexlab{}.
\newblock \showarticletitle{Making group decisions from natural language-based preferences}. In \bibinfo{booktitle}{\emph{Proceedings of the 8th International Workshop on Computational Social Choice (COMSOC)}}. \bibinfo{pages}{2}.
\newblock


\bibitem[Monarch(2021)]%
        {monarch2021human}
\bibfield{author}{\bibinfo{person}{Robert~Munro Monarch}.} \bibinfo{year}{2021}\natexlab{}.
\newblock \bibinfo{booktitle}{\emph{Human-in-the-Loop Machine Learning: Active learning and annotation for human-centered AI}}.
\newblock \bibinfo{publisher}{Simon and Schuster}.
\newblock


\bibitem[Mosqueira-Rey et~al\mbox{.}(2023)]%
        {mosqueira-rey_human---loop_2023}
\bibfield{author}{\bibinfo{person}{Eduardo Mosqueira-Rey}, \bibinfo{person}{Elena Hernández-Pereira}, \bibinfo{person}{David Alonso-Ríos}, \bibinfo{person}{José Bobes-Bascarán}, {and} \bibinfo{person}{Ángel Fernández-Leal}.} \bibinfo{year}{2023}\natexlab{}.
\newblock \showarticletitle{Human-in-the-loop machine learning: a state of the art}.
\newblock \bibinfo{journal}{\emph{Artificial Intelligence Review}} \bibinfo{volume}{56}, \bibinfo{number}{4} (\bibinfo{date}{April} \bibinfo{year}{2023}), \bibinfo{pages}{3005--3054}.
\newblock
\showISSN{1573-7462}
\urldef\tempurl%
\url{https://doi.org/10.1007/s10462-022-10246-w}
\showDOI{\tempurl}


\bibitem[Neu and Szepesv{\'a}ri(2009)]%
        {neu2009training}
\bibfield{author}{\bibinfo{person}{Gergely Neu} {and} \bibinfo{person}{Csaba Szepesv{\'a}ri}.} \bibinfo{year}{2009}\natexlab{}.
\newblock \showarticletitle{Training parsers by inverse reinforcement learning}.
\newblock \bibinfo{journal}{\emph{Machine learning}}  \bibinfo{volume}{77} (\bibinfo{year}{2009}), \bibinfo{pages}{303--337}.
\newblock


\bibitem[Ng et~al\mbox{.}(2000)]%
        {ng2000algorithms}
\bibfield{author}{\bibinfo{person}{Andrew~Y Ng}, \bibinfo{person}{Stuart Russell}, {et~al\mbox{.}}} \bibinfo{year}{2000}\natexlab{}.
\newblock \showarticletitle{Algorithms for inverse reinforcement learning.}. In \bibinfo{booktitle}{\emph{Icml}}, Vol.~\bibinfo{volume}{1}. \bibinfo{pages}{2}.
\newblock


\bibitem[Nguyen and Smeulders(2004)]%
        {nguyen2004active}
\bibfield{author}{\bibinfo{person}{Hieu~T Nguyen} {and} \bibinfo{person}{Arnold Smeulders}.} \bibinfo{year}{2004}\natexlab{}.
\newblock \showarticletitle{Active learning using pre-clustering}. In \bibinfo{booktitle}{\emph{Proceedings of the twenty-first international conference on Machine learning}}. \bibinfo{pages}{79}.
\newblock


\bibitem[Noothigattu et~al\mbox{.}(2018)]%
        {noothigattu2018voting}
\bibfield{author}{\bibinfo{person}{Ritesh Noothigattu}, \bibinfo{person}{Snehalkumar Gaikwad}, \bibinfo{person}{Edmond Awad}, \bibinfo{person}{Sohan Dsouza}, \bibinfo{person}{Iyad Rahwan}, \bibinfo{person}{Pradeep Ravikumar}, {and} \bibinfo{person}{Ariel Procaccia}.} \bibinfo{year}{2018}\natexlab{}.
\newblock \showarticletitle{A voting-based system for ethical decision making}. In \bibinfo{booktitle}{\emph{Proceedings of the AAAI Conference on Artificial Intelligence}}, Vol.~\bibinfo{volume}{32}.
\newblock


\bibitem[Olsson(2009)]%
        {olsson2009literature}
\bibfield{author}{\bibinfo{person}{Fredrik Olsson}.} \bibinfo{year}{2009}\natexlab{}.
\newblock \showarticletitle{A literature survey of active machine learning in the context of natural language processing}.
\newblock  (\bibinfo{year}{2009}).
\newblock


\bibitem[Pardo et~al\mbox{.}(2018)]%
        {pardo2018time}
\bibfield{author}{\bibinfo{person}{Fabio Pardo}, \bibinfo{person}{Arash Tavakoli}, \bibinfo{person}{Vitaly Levdik}, {and} \bibinfo{person}{Petar Kormushev}.} \bibinfo{year}{2018}\natexlab{}.
\newblock \showarticletitle{Time limits in reinforcement learning}. In \bibinfo{booktitle}{\emph{International Conference on Machine Learning}}. PMLR, \bibinfo{pages}{4045--4054}.
\newblock


\bibitem[Price et~al\mbox{.}(2003)]%
        {price2003new}
\bibfield{author}{\bibinfo{person}{Sara Price}, \bibinfo{person}{Yvonne Rogers}, \bibinfo{person}{Danae Stanton}, {and} \bibinfo{person}{Hilary Smith}.} \bibinfo{year}{2003}\natexlab{}.
\newblock \showarticletitle{A new conceptual framework for CSCL: Supporting diverse forms of reflection through multiple interactions}. In \bibinfo{booktitle}{\emph{Designing for change in networked learning environments: Proceedings of the International Conference on Computer Support for Collaborative Learning 2003}}. Springer, \bibinfo{pages}{513--522}.
\newblock


\bibitem[Priyogi(2019)]%
        {priyogi_preference_2019}
\bibfield{author}{\bibinfo{person}{Bilih Priyogi}.} \bibinfo{year}{2019}\natexlab{}.
\newblock \showarticletitle{Preference {Elicitation} {Strategy} for {Conversational} {Recommender} {System}}. In \bibinfo{booktitle}{\emph{Proceedings of the {Twelfth} {ACM} {International} {Conference} on {Web} {Search} and {Data} {Mining}}}. \bibinfo{publisher}{ACM}, \bibinfo{address}{Melbourne VIC Australia}, \bibinfo{pages}{824--825}.
\newblock
\showISBNx{978-1-4503-5940-5}
\urldef\tempurl%
\url{https://doi.org/10.1145/3289600.3291604}
\showDOI{\tempurl}


\bibitem[Radlinski et~al\mbox{.}(2019)]%
        {radlinski_coached_2019}
\bibfield{author}{\bibinfo{person}{Filip Radlinski}, \bibinfo{person}{Krisztian Balog}, \bibinfo{person}{Bill Byrne}, {and} \bibinfo{person}{Karthik Krishnamoorthi}.} \bibinfo{year}{2019}\natexlab{}.
\newblock \showarticletitle{Coached {Conversational} {Preference} {Elicitation}: {A} {Case} {Study} in {Understanding} {Movie} {Preferences}}. In \bibinfo{booktitle}{\emph{Proceedings of the 20th {Annual} {SIGdial} {Meeting} on {Discourse} and {Dialogue}}}, \bibfield{editor}{\bibinfo{person}{Satoshi Nakamura}, \bibinfo{person}{Milica Gasic}, \bibinfo{person}{Ingrid Zukerman}, \bibinfo{person}{Gabriel Skantze}, \bibinfo{person}{Mikio Nakano}, \bibinfo{person}{Alexandros Papangelis}, \bibinfo{person}{Stefan Ultes}, {and} \bibinfo{person}{Koichiro Yoshino}} (Eds.). \bibinfo{publisher}{Association for Computational Linguistics}, \bibinfo{address}{Stockholm, Sweden}, \bibinfo{pages}{353--360}.
\newblock
\urldef\tempurl%
\url{https://doi.org/10.18653/v1/W19-5941}
\showDOI{\tempurl}


\bibitem[Ramachandran and Amir(2007)]%
        {ramachandran2007bayesian}
\bibfield{author}{\bibinfo{person}{Deepak Ramachandran} {and} \bibinfo{person}{Eyal Amir}.} \bibinfo{year}{2007}\natexlab{}.
\newblock \showarticletitle{Bayesian Inverse Reinforcement Learning.}. In \bibinfo{booktitle}{\emph{IJCAI}}, Vol.~\bibinfo{volume}{7}. \bibinfo{pages}{2586--2591}.
\newblock


\bibitem[Ren et~al\mbox{.}(2021)]%
        {ren2021survey}
\bibfield{author}{\bibinfo{person}{Pengzhen Ren}, \bibinfo{person}{Yun Xiao}, \bibinfo{person}{Xiaojun Chang}, \bibinfo{person}{Po-Yao Huang}, \bibinfo{person}{Zhihui Li}, \bibinfo{person}{Brij~B Gupta}, \bibinfo{person}{Xiaojiang Chen}, {and} \bibinfo{person}{Xin Wang}.} \bibinfo{year}{2021}\natexlab{}.
\newblock \showarticletitle{A survey of deep active learning}.
\newblock \bibinfo{journal}{\emph{ACM computing surveys (CSUR)}} \bibinfo{volume}{54}, \bibinfo{number}{9} (\bibinfo{year}{2021}), \bibinfo{pages}{1--40}.
\newblock


\bibitem[Renner et~al\mbox{.}(2016)]%
        {renner2016effects}
\bibfield{author}{\bibinfo{person}{Bettina Renner}, \bibinfo{person}{Michael Prilla}, \bibinfo{person}{Ulrike Cress}, {and} \bibinfo{person}{Joachim Kimmerle}.} \bibinfo{year}{2016}\natexlab{}.
\newblock \showarticletitle{Effects of prompting in reflective learning tools: Findings from experimental field, lab, and online studies}.
\newblock \bibinfo{journal}{\emph{Frontiers in psychology}}  \bibinfo{volume}{7} (\bibinfo{year}{2016}), \bibinfo{pages}{820}.
\newblock


\bibitem[Ribeiro(1996)]%
        {ribeiro1996fuzzy}
\bibfield{author}{\bibinfo{person}{Rita~Almeida Ribeiro}.} \bibinfo{year}{1996}\natexlab{}.
\newblock \showarticletitle{Fuzzy multiple attribute decision making: a review and new preference elicitation techniques}.
\newblock \bibinfo{journal}{\emph{Fuzzy sets and systems}} \bibinfo{volume}{78}, \bibinfo{number}{2} (\bibinfo{year}{1996}), \bibinfo{pages}{155--181}.
\newblock


\bibitem[Rogers and Muller(2006)]%
        {rogers2006framework}
\bibfield{author}{\bibinfo{person}{Yvonne Rogers} {and} \bibinfo{person}{Henk Muller}.} \bibinfo{year}{2006}\natexlab{}.
\newblock \showarticletitle{A framework for designing sensor-based interactions to promote exploration and reflection in play}.
\newblock \bibinfo{journal}{\emph{International Journal of Human-Computer Studies}} \bibinfo{volume}{64}, \bibinfo{number}{1} (\bibinfo{year}{2006}), \bibinfo{pages}{1--14}.
\newblock


\bibitem[Russell(2019)]%
        {russell2019human}
\bibfield{author}{\bibinfo{person}{Stuart Russell}.} \bibinfo{year}{2019}\natexlab{}.
\newblock \bibinfo{booktitle}{\emph{Human compatible: AI and the problem of control}}.
\newblock \bibinfo{publisher}{Penguin Uk}.
\newblock


\bibitem[Sarwar et~al\mbox{.}(2001)]%
        {sarwar2001item}
\bibfield{author}{\bibinfo{person}{Badrul Sarwar}, \bibinfo{person}{George Karypis}, \bibinfo{person}{Joseph Konstan}, {and} \bibinfo{person}{John Riedl}.} \bibinfo{year}{2001}\natexlab{}.
\newblock \showarticletitle{Item-based collaborative filtering recommendation algorithms}. In \bibinfo{booktitle}{\emph{Proceedings of the 10th international conference on World Wide Web}}. \bibinfo{pages}{285--295}.
\newblock


\bibitem[Schwartz(1992)]%
        {schwartz1992universals}
\bibfield{author}{\bibinfo{person}{Shalom~H Schwartz}.} \bibinfo{year}{1992}\natexlab{}.
\newblock \showarticletitle{Universals in the content and structure of values: Theoretical advances and empirical tests in 20 countries}.
\newblock \bibinfo{journal}{\emph{Advances in experimental social psychology/Academic Press}} (\bibinfo{year}{1992}).
\newblock


\bibitem[Settles(2009)]%
        {settles2009active}
\bibfield{author}{\bibinfo{person}{Burr Settles}.} \bibinfo{year}{2009}\natexlab{}.
\newblock \showarticletitle{Active learning literature survey}.
\newblock  (\bibinfo{year}{2009}).
\newblock


\bibitem[Siththaranjan et~al\mbox{.}(2023)]%
        {siththaranjan2023distributional}
\bibfield{author}{\bibinfo{person}{Anand Siththaranjan}, \bibinfo{person}{Cassidy Laidlaw}, {and} \bibinfo{person}{Dylan Hadfield-Menell}.} \bibinfo{year}{2023}\natexlab{}.
\newblock \showarticletitle{Understanding Hidden Context in Preference Learning: Consequences for RLHF}. In \bibinfo{booktitle}{\emph{The Twelfth International Conference on Learning Representations}}.
\newblock


\bibitem[Sun and Zhang(2018)]%
        {sun2018conversational}
\bibfield{author}{\bibinfo{person}{Yueming Sun} {and} \bibinfo{person}{Yi Zhang}.} \bibinfo{year}{2018}\natexlab{}.
\newblock \showarticletitle{Conversational recommender system}. In \bibinfo{booktitle}{\emph{The 41st international acm sigir conference on research \& development in information retrieval}}. \bibinfo{pages}{235--244}.
\newblock


\bibitem[Van~de Poel(2013)]%
        {van2013translating}
\bibfield{author}{\bibinfo{person}{Ibo Van~de Poel}.} \bibinfo{year}{2013}\natexlab{}.
\newblock \showarticletitle{Translating values into design requirements}.
\newblock \bibinfo{journal}{\emph{Philosophy and engineering: Reflections on practice, principles and process}} (\bibinfo{year}{2013}), \bibinfo{pages}{253--266}.
\newblock


\bibitem[Viappiani et~al\mbox{.}(2006)]%
        {viappiani2006preference}
\bibfield{author}{\bibinfo{person}{Paolo Viappiani}, \bibinfo{person}{Boi Faltings}, {and} \bibinfo{person}{Pearl Pu}.} \bibinfo{year}{2006}\natexlab{}.
\newblock \showarticletitle{Preference-based search using example-critiquing with suggestions}.
\newblock \bibinfo{journal}{\emph{Journal of artificial intelligence Research}}  \bibinfo{volume}{27} (\bibinfo{year}{2006}), \bibinfo{pages}{465--503}.
\newblock


\bibitem[Williams et~al\mbox{.}(2016)]%
        {williams2016revising}
\bibfield{author}{\bibinfo{person}{Joseph~Jay Williams}, \bibinfo{person}{Tania Lombrozo}, \bibinfo{person}{Anne Hsu}, \bibinfo{person}{Bernd Huber}, {and} \bibinfo{person}{Juho Kim}.} \bibinfo{year}{2016}\natexlab{}.
\newblock \showarticletitle{Revising learner misconceptions without feedback: Prompting for reflection on anomalies}. In \bibinfo{booktitle}{\emph{Proceedings of the 2016 CHI conference on human factors in computing systems}}. \bibinfo{pages}{470--474}.
\newblock


\bibitem[Wolfbauer et~al\mbox{.}(2022a)]%
        {wolfbauer_script_2022}
\bibfield{author}{\bibinfo{person}{Irmtraud Wolfbauer}, \bibinfo{person}{Viktoria Pammer-Schindler}, \bibinfo{person}{Katharina Maitz}, {and} \bibinfo{person}{Carolyn~P. Rose}.} \bibinfo{year}{2022}\natexlab{a}.
\newblock \showarticletitle{A {Script} for {Conversational} {Reflection} {Guidance}: {A} {Field} {Study} on {Developing} {Reflection} {Competence} {With} {Apprentices}}.
\newblock \bibinfo{journal}{\emph{IEEE Transactions on Learning Technologies}} \bibinfo{volume}{15}, \bibinfo{number}{5} (\bibinfo{date}{Oct.} \bibinfo{year}{2022}), \bibinfo{pages}{554--566}.
\newblock
\showISSN{1939-1382, 2372-0050}
\urldef\tempurl%
\url{https://doi.org/10.1109/TLT.2022.3207226}
\showDOI{\tempurl}


\bibitem[Wolfbauer et~al\mbox{.}(2022b)]%
        {wolfbauer2022script}
\bibfield{author}{\bibinfo{person}{Irmtraud Wolfbauer}, \bibinfo{person}{Viktoria Pammer-Schindler}, \bibinfo{person}{Katharina Maitz}, {and} \bibinfo{person}{Carolyn~P Ros{\'e}}.} \bibinfo{year}{2022}\natexlab{b}.
\newblock \showarticletitle{A script for conversational reflection guidance: a field study on developing reflection competence with apprentices}.
\newblock \bibinfo{journal}{\emph{IEEE Transactions on Learning Technologies}} \bibinfo{volume}{15}, \bibinfo{number}{5} (\bibinfo{year}{2022}), \bibinfo{pages}{554--566}.
\newblock


\bibitem[Yin et~al\mbox{.}(2024)]%
        {yin_jamplate_2024}
\bibfield{author}{\bibinfo{person}{Jiayu Yin}, \bibinfo{person}{Catherine Gu}, \bibinfo{person}{Jenny Mar}, {and} \bibinfo{person}{Sydney Zhang}.} \bibinfo{year}{2024}\natexlab{}.
\newblock \showarticletitle{Jamplate: {Exploring} {LLM}-{Enhanced} {Templates} for {Idea} {Reflection}}.
\newblock  (\bibinfo{year}{2024}).
\newblock


\bibitem[Yukawa(2003)]%
        {yukawa2003co}
\bibfield{author}{\bibinfo{person}{Joyce Yukawa}.} \bibinfo{year}{2003}\natexlab{}.
\newblock \showarticletitle{Co-reflection in online learning environments}.
\newblock \bibinfo{journal}{\emph{ACM SIGGROUP Bulletin}} \bibinfo{volume}{24}, \bibinfo{number}{3} (\bibinfo{year}{2003}), \bibinfo{pages}{44--49}.
\newblock


\bibitem[Ziebart et~al\mbox{.}(2008)]%
        {ziebart2008maximum}
\bibfield{author}{\bibinfo{person}{Brian~D Ziebart}, \bibinfo{person}{Andrew~L Maas}, \bibinfo{person}{J~Andrew Bagnell}, \bibinfo{person}{Anind~K Dey}, {et~al\mbox{.}}} \bibinfo{year}{2008}\natexlab{}.
\newblock \showarticletitle{Maximum entropy inverse reinforcement learning.}. In \bibinfo{booktitle}{\emph{Aaai}}, Vol.~\bibinfo{volume}{8}. Chicago, IL, USA, \bibinfo{pages}{1433--1438}.
\newblock


\bibitem[Ziegler et~al\mbox{.}(2019)]%
        {ziegler2019fine}
\bibfield{author}{\bibinfo{person}{Daniel~M Ziegler}, \bibinfo{person}{Nisan Stiennon}, \bibinfo{person}{Jeffrey Wu}, \bibinfo{person}{Tom~B Brown}, \bibinfo{person}{Alec Radford}, \bibinfo{person}{Dario Amodei}, \bibinfo{person}{Paul Christiano}, {and} \bibinfo{person}{Geoffrey Irving}.} \bibinfo{year}{2019}\natexlab{}.
\newblock \showarticletitle{Fine-tuning language models from human preferences}.
\newblock \bibinfo{journal}{\emph{arXiv preprint arXiv:1909.08593}} (\bibinfo{year}{2019}).
\newblock


\end{thebibliography}

% \appendix
\clearpage
\appendix

\section{System Algorithm}

\RestyleAlgo{ruled}
\begin{algorithm}[!h]
\small
\DontPrintSemicolon
\caption{\systemname{} }
\label{alg:IRDA}

\KwData{
Desired behaviour \texttt{lang(behaviour)}, number of diversity-based trajectories to query the user about $k$, confidence threshold $\epsilon$
}
% \KwResult{Personalized reward model}

\SetKwBlock{Begin}{function}{end function}
% \Begin(){
    Diversity-based sampling of \( k \) trajectories using k-means\;
    \For{\(i \) in \( k \)}{
        Show video of trajectory $i$\;
        Collect user feedback\;
        Add ASCII encoding of trajectory and user feedback to reward model\;
    }
    Generate user feature hypothesis and alternative features\;
    Collect user reflection\;
    Add hypothesis, alternative features, and user reflection to the reward model.
    
    \If{\text{user chooses to enter the preference clarification loop}}{
        return to step 2\;
    }
    \While{model confidence $< \epsilon$}{
        Use uncertainty sampling to select trajectory\;
        Show video of trajectory\;
        Collect user feedback\;
        Add ASCII encoding of trajectory and feedback to reward model\;
        Recalculate \textit{model confidence}   
    }
    \KwRet{Reward model}
% }
\end{algorithm}

\section{Trajectory Encoding}\label{appendix:trajectory_encoding}
\subsection{Study 1 - Multi-Agent Apple Farming}
During the feedback collection phase, the user converses with the system about how they would like the agent to act. All of the feedback collected from the user is in natural language. Thus, the reward model must be able to make use of natural language information. However, the information about the agent and environment may not be in natural language. For example, in the multi-agent apple farming environment, the information about the environment and the agent's actions is stored in a numerical format that can later be rendered as an RGB array when a human wants to view it. This presents a challenge as the feedback from the user and the information about what the feedback is referring to, namely the agent and the environment, are not in the same format.

One possible solution to this grounding problem is to use a vision-language model (VLM) that can accept both natural language and visual information. However, when we performed preliminary testing, we found that the performance of VLMs was notably poor when attempting to prompt the model to reason about detailed spatial information from reinforcement learning environments such as grid worlds. It is possible that as VLMs improve, they will be able to reason about high-granularity spatial images successfully. Regardless, given current capabilities, we opted to convert the trajectories into ASCII representations, which an LLM can understand.

To do so, we coded each entity in the environment with an ASCII character. For example, for the environment used in our evaluation, the main agent the user was giving feedback on was represented by an ``\texttt{M}'', apples, which the agent could collect for reward, by ``\texttt{A}'', other agents by ``\texttt{B}'', garbage by ``\texttt{G}'', empty tiles by ``\texttt{.}'', and orchard boundaries by ``\texttt{-}'' and ``\texttt{|}''. We also annotated each step in the trajectory with the step number and important actions that occurred. Figure \ref{fig:ascii-encoding} shows an example of two timesteps encoded in this way. Notably, all of the information comes from the underlying state of the environment and is programmatically generated. This means a human does not have to convert and annotate each trajectory, which is important for scalability.

\subsection{Study 2 - The Moral Machine}
\subsubsection{Numerical}
The Moral Machine dataset is transformed into a 26-dimensional vector representation for each scenario by selecting relevant features from the original dataset described in detail here: \textcolor{blue}{\href{https://osf.io/wt6mc}{https://osf.io/wt6mc}}. These features include scenario structure elements (Intervention, PedPed, Barrier, CrossingSignal), numeric characteristics (NumberOfCharacters, DiffNumberOFCharacters), and 20 character types (e.g., Man, Woman, Pregnant, etc.). Each moral dilemma compares two outcomes (stay vs. swerve), represented by two rows in the dataset. The vectorization process involves creating comparison pairs based on the 'Intervention' column, then calculating the difference between feature values of the two outcomes. This differencing captures the relative changes in features between outcomes, resulting in a 26-dimensional difference vector. Before use in the Multi-Layer Perceptron classifier, these vectors are standardized using StandardScaler to ensure consistent feature scaling.

\subsubsection{ASCII}
\begin{figure}[H]
    \centering
    \begin{minipage}{0.4\textwidth}
    \begin{lstlisting}
The brakes of a self-driving car have 
failed. The self-driving car can 
continue driving straight ahead 
or swerve. If the car continues
straight ahead, it will run over
a group of 5 pedestrians who are 
crossing against a red do not walk 
signal. The group of pedestrians 
include:
    - 4 girls
    - A female doctor
If the car swerves, it will run over a 
group of 5 pedestrians who are crossing
with a green walk signal. The group of 
pedestrians include:
    - 4 boys
    - A male doctor
    \end{lstlisting}
    \end{minipage}
    \caption{ASCII encoding of a Moral Machine scenario from Study 2.}
    \label{fig:ascii-encoding_s2}
\end{figure}

The process of creating natural language descriptions for the Moral Machine scenarios involves converting the raw data into a verbal description. 
Each scenario is described by presenting the basic dilemma of a self-driving car with failed brakes, followed by the two possible outcomes (continuing straight or swerving). 
The description includes details about the number and types of characters involved in each outcome, their actions (such as crossing legally or illegally), and any relevant attributes (like profession or age). 
An example is shown in \autoref{fig:ascii-encoding_s2}.

\section{Detailed Description of Supervised Learning Baselines}\label{appendix:SL}
The MLP models consist of one hidden layer with 32 neurons, trained with a learning rate of 0.001 and the Adam optimizer. For Study 1, the input was based on the grid map encoding, while for Study 2, a 26-dimension vector representing Moral Machine scenarios was used.
The CNN architecture for Study 2 comprises two convolutional layers followed by ReLU activation and max pooling. The first layer has 16 filters, and the second has 32 filters, both with a kernel size of 3 and padding of 1. After flattening, there are two fully connected layers with ReLU activation, reducing the dimensionality to 64 and then to the final output size of 2.
These architectures were chosen to balance model complexity with the limited amount of training data available per participant. The use of both individual and collective models allows us to compare personalized preferences with aggregated group preferences across different neural network structures and input representations.

\section{Jaccard Similarity Visualizations}

\begin{figure}[H]
  \centering
  \includegraphics[width=0.98\linewidth]{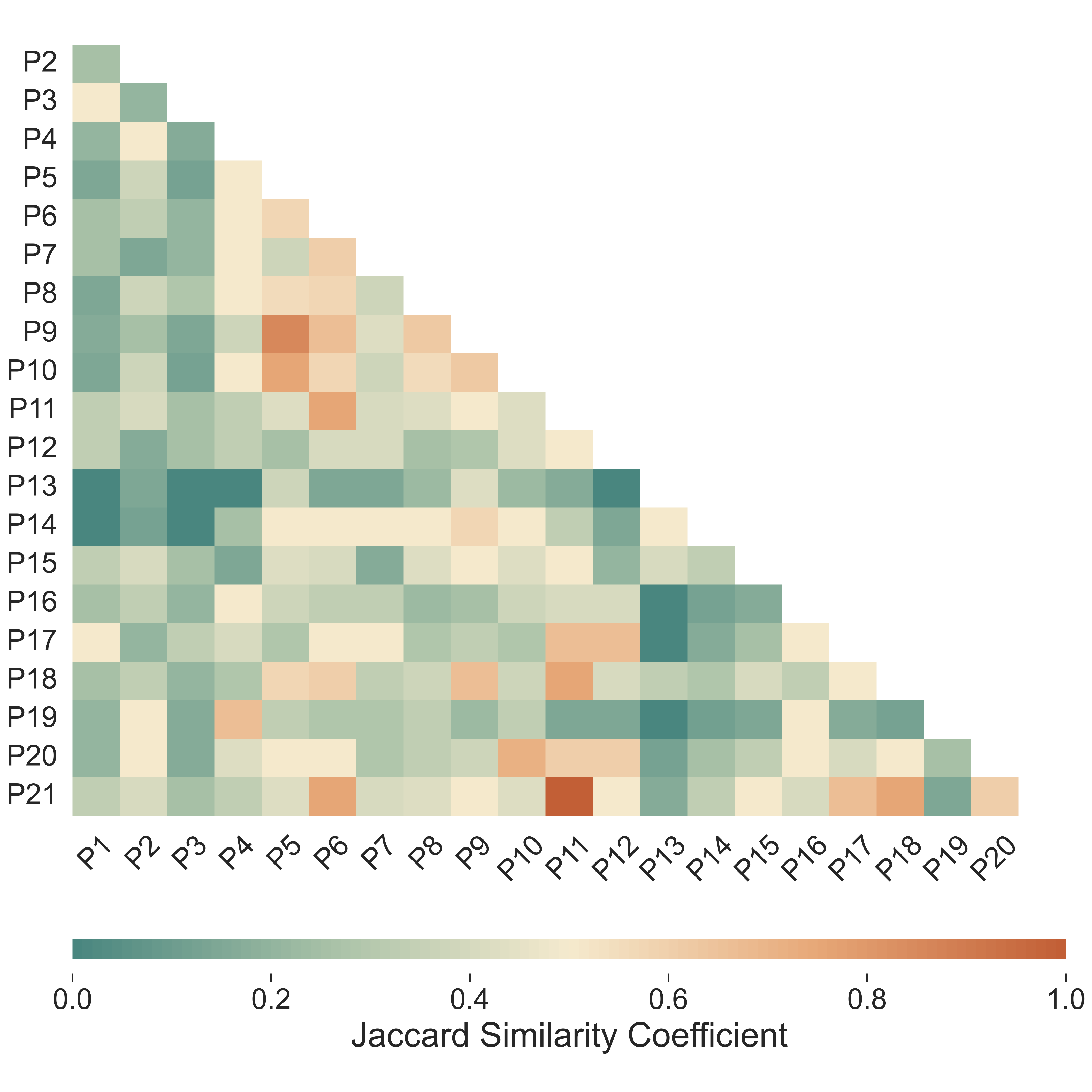}
  \caption{Heatmap of pairwise Jaccard similarity coefficients between participants (P1-P21) based on their use of decision-making features in Study 1. The Jaccard similarity coefficient quantifies the overlap in features used by each pair of participants, with values ranging from 0 (no overlap) to 1 (complete overlap).}
  \Description{Heatmap of Jaccard Similarity Coefficients between participants' (P1-P21) feature considerations in Study 1. The triangular matrix shows pairwise similarities, with darker green indicating higher similarity and darker orange lower similarity. Most pairs exhibit moderate similarity (light green to beige), suggesting some consensus in feature consideration. However, the presence of both very light and very dark cells indicates notable variations in approach among some participants. Diagonal elements are omitted as they would always show perfect similarity. This visualization helps identify clusters of participants with similar decision-making strategies and highlights outliers with unique approaches.}
  \label{jaccard_s1}
\end{figure}

\begin{figure}[H]
  \centering
  \includegraphics[width=0.98\linewidth]{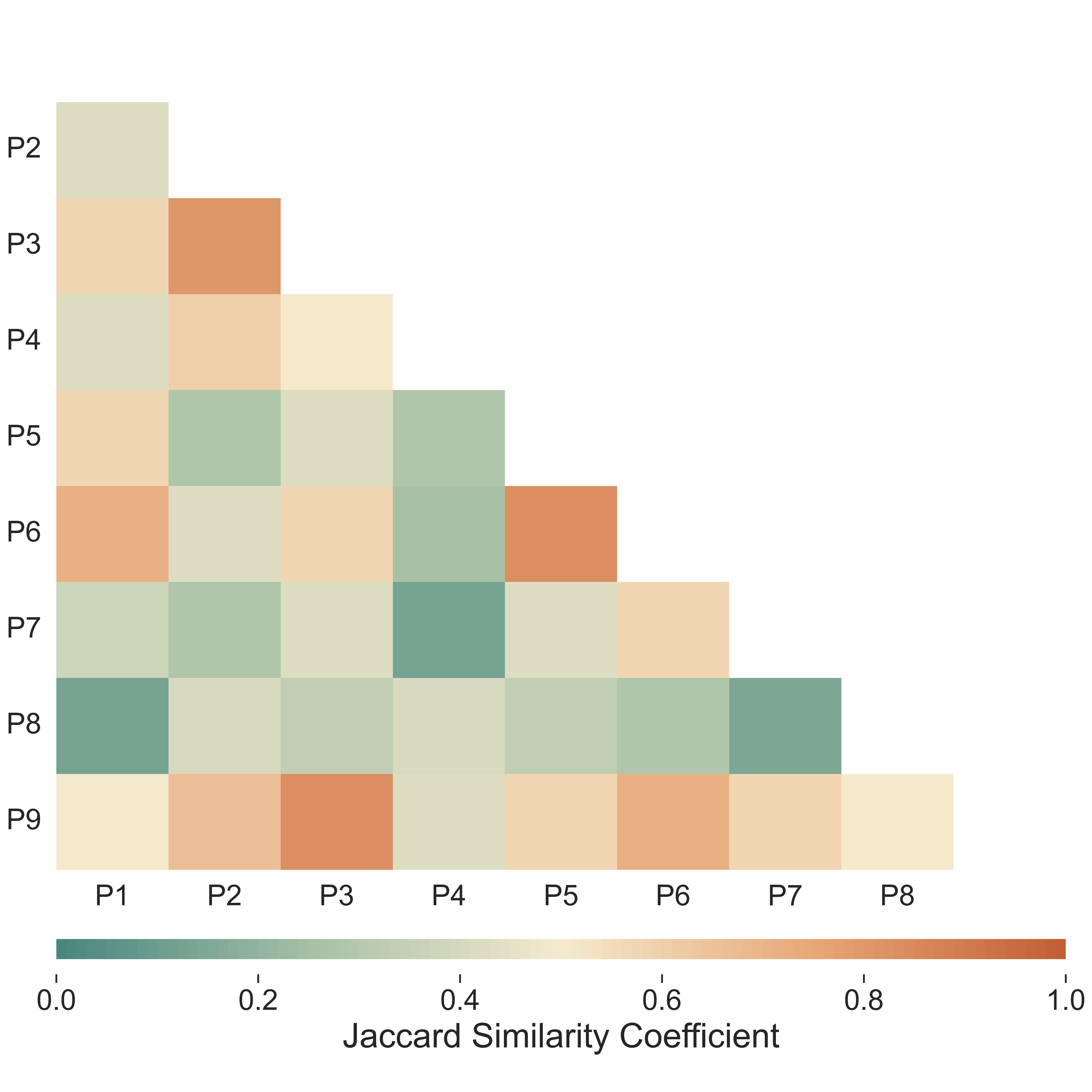}
  \caption{Heatmap of pairwise Jaccard similarity coefficients between participants (P1-P9) based on their use of decision-making features in Study 2. The Jaccard similarity coefficient quantifies the overlap in features used by each pair of participants, with values ranging from 0 (no overlap) to 1 (complete overlap).}
  \Description{Heatmap of pairwise Jaccard similarity coefficients between participants (P1-P9) based on their consideration of decision-making features in Study 2. The triangular matrix shows varying degrees of similarity, with darker green indicating lower similarity and darker orange higher similarity. Most pairs exhibit moderate similarity (light green to light orange), suggesting some consensus in feature consideration. However, the presence of both very light and very dark cells indicates notable variations among some participants. For instance, P1 and P8 show high similarity (dark green), while P2 and P3 have low similarity (orange).}
  \label{jaccard_s2}
\end{figure}

\end{document}